\documentclass[11pt]{article}

\usepackage[preprint]{acl}

\usepackage{silence}
\usepackage{microtype}
\usepackage{listings}
\usepackage{comment}

\usepackage{amsmath}
\usepackage{amssymb}
\usepackage{xcolor}
\usepackage{float}
\usepackage[nomarkers,notablist,notabhead]{endfloat}
\usepackage{tabularx}
\usepackage{longtable}
\usepackage{booktabs}
\usepackage{multirow}
\usepackage{algorithm}
\usepackage{algpseudocode}
\usepackage{graphicx}
\graphicspath{{./}{../}}
\usepackage{pgfplots}
\usepackage[nameinlink]{cleveref}
\usepackage{enumitem}
\usepackage[normalem]{ulem}

\pgfplotsset{compat=1.18}
\makeatletter
\let\AppendixDelayedTable\table
\expandafter\let\expandafter\AppendixDelayedTableStar\csname table*\endcsname
\AtBeginDocument{\efloat@restorefloats}
\makeatother

\usepackage{fontspec}
\usepackage[english,bidi=default]{babel} 
\babelfont{rm}{TeXGyreTermesX} 

\babelprovide[import]{arabic}
\IfFontExistsTF{Noto Sans Arabic}{\babelfont[*arabic]{rm}{Noto Sans Arabic}}{}

\babelprovide[import]{armenian}   
\babelprovide[import]{belarusian} 
\babelprovide[import]{spanish}    
\babelprovide[import]{indonesian} 
\babelprovide[import]{khmer}      
\babelprovide[import]{lao}        
\babelprovide[import]{burmese}    
\babelprovide[import]{tamil}      
\babelprovide[import]{thai}       
\babelprovide[import]{chinese}    

\IfFontExistsTF{Noto Sans Khmer}{\babelfont[khmer]{rm}{Noto Sans Khmer}}{}
\IfFontExistsTF{Noto Sans Lao}{\babelfont[lao]{rm}{Noto Sans Lao}}{}
\IfFontExistsTF{Noto Serif Armenian}{\babelfont[armenian]{rm}{Noto Serif Armenian}}{}
\IfFontExistsTF{Noto Sans Armenian}{\babelfont[armenian]{rm}{Noto Sans Armenian}}{}
\newfontfamily\amharicfont[
    Path=fonts/,
    Extension=.ttf
]{NotoSansEthiopic-Regular}
\IfFontExistsTF{Noto Sans}{\babelfont[belarusian]{rm}{Noto Sans}}{}
\IfFontExistsTF{Noto Sans}{\babelfont[spanish]{rm}{Noto Sans}}{}
\IfFontExistsTF{Noto Sans Myanmar}{\babelfont[burmese]{rm}{Noto Sans Myanmar}}{}
\IfFontExistsTF{Noto Sans Tamil}{\babelfont[tamil]{rm}{Noto Sans Tamil}}{}
\IfFontExistsTF{Noto Sans Thai}{\babelfont[thai]{rm}{Noto Sans Thai}}{}
\IfFontExistsTF{Noto Sans Tagalog}{\babelfont[filipino]{rm}{Noto Sans Tagalog}}{}
\babelfont[chinese]{rm}[
    Path = fonts/,
    UprightFont = NotoSansCJKsc-Regular.otf,
    BoldFont = NotoSansCJKsc-Bold.otf,
    AutoFakeSlant = 0.15
]{NotoSansCJKsc-Regular.otf}

\title{Parameter-Efficient Vocabulary Expansion \\ via Cross-Lingual Embedding Alignment}

\author{
  \textbf{Andrew Ivan Soegeng}$^{1,2}$, \textbf{Muhammad Reza Qorib}$^{3}$, \textbf{Weishi Wang}$^{2}$, \\
  \textbf{Daniel Dahlmeier}$^{2}$, \textbf{Hwee Tou Ng}$^{1}$ \\
  $^{1}$National University of Singapore, $^{2}$SAP, $^{3}$Carnegie Mellon University \\
  \texttt{\{andrew.soegeng,weishi.wang,d.dahlmeier\}@sap.com} \\
  \texttt{mrqorib@cmu.edu} \\
  \texttt{dcsnght@nus.edu.sg}
}
\begin{document}

\maketitle
\begin{abstract}
Continual pretraining with vocabulary expansion is often used to make large language models (LLMs) more effective and efficient at handling new languages. However, when LLMs are extended to many languages simultaneously, the number of model parameters increases significantly. In addition, the newly added tokens do not receive the same number of training steps as the tokens in the original vocabulary, which sometimes causes vocabulary expansion to yield only limited performance improvements. We propose leveraging the semantic equivalence between words in the model’s primary language and those in the extended vocabulary to transfer learned representations. In this paper, we introduce \textsc{MATE}, a parameter-efficient vocabulary expansion method that enables effective cross-lingual transfer with 8× fewer additional parameters per token. \textsc{MATE} reduces the average tokenization compression ratio and delivers improvements on multiple benchmarks.
\end{abstract}

\section{Introduction}

Large Language Models (LLMs) are predominantly pre-trained on English and a few high-resource languages \cite{brown_language_2020, touvron_llama_2023, olmo_2_2025}. Adapting them to unseen, low-resource languages is often bottlenecked by static subword tokenizers, e.g., BPE \cite{sennrich_neural_2016}. When applied to new scripts, these tokenizers tend to fragment words inefficiently into byte-level sequences, resulting in high token fertility, increased latency, and diluted semantics \cite{rust_how_2021}. For example, OLMo 2 \cite{olmo_2_2025} tokenizes a Burmese text into a sequence 12 times longer than its English equivalent.

\begin{figure}
    \centering
    \includegraphics[width=1\linewidth]{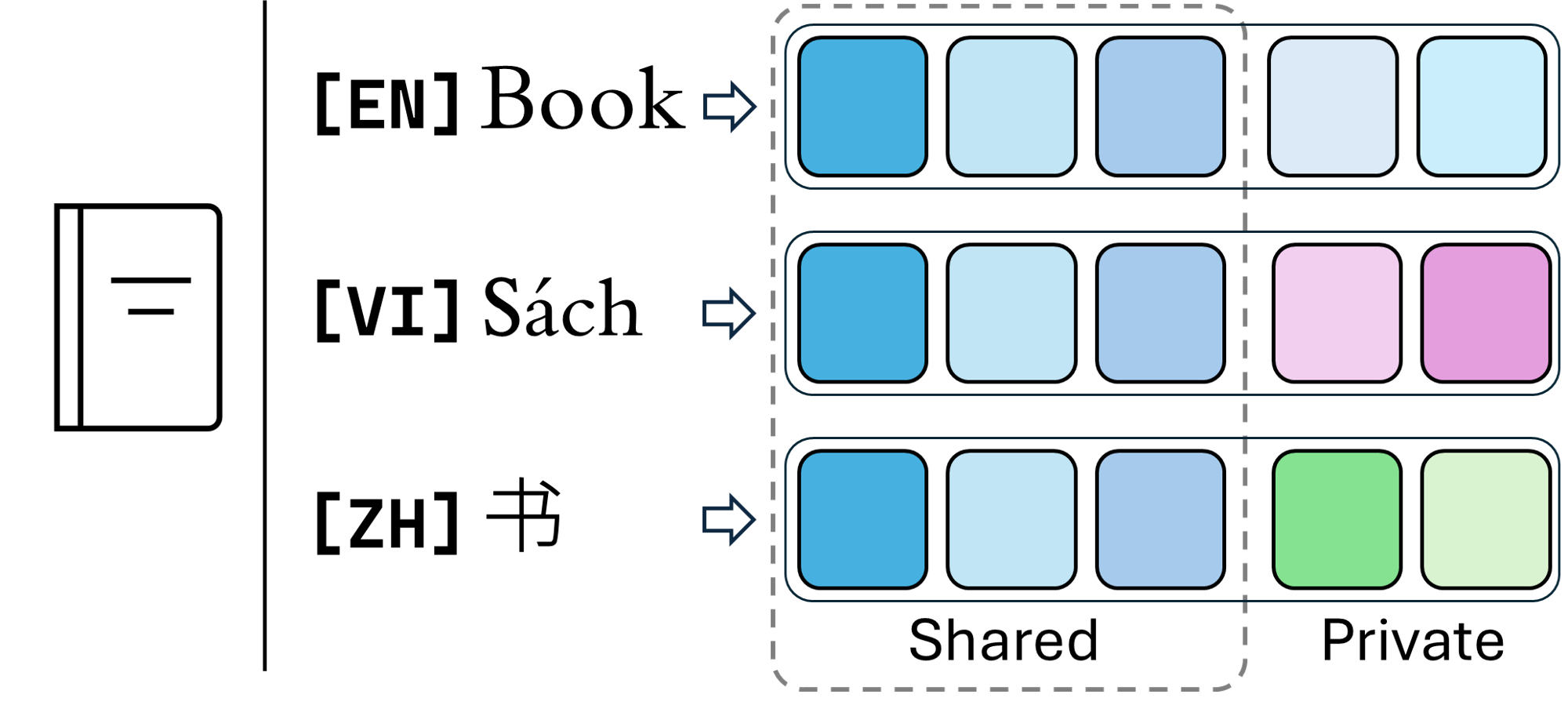}
    \caption{Illustration of embedding vectors for words with the same meaning across different languages using MATE. With English as the anchor language, some embedding dimensions in the other languages are weight-tied to improve cross-lingual transfer, while learnable private parameters capture language-specific nuances.}
    \label{fig:mshare_overview}
\end{figure}

Prevailing solutions rely on vocabulary adaptation, such as replacing or expanding the tokenizer \cite{li_tokalign_2025, fujii_continual_2024, cui_efficient_2023}. However, these methods have several drawbacks. First, they increase the number of model parameters. More importantly, standard vocabulary expansion, which initializes new target-language tokens as independent parameters, often fails to improve downstream performance \cite{fujii_continual_2024}. This may occur because the new tokens are undertrained, as they receive far fewer training steps than the original vocabulary. In addition, such practices can also lead to catastrophic forgetting~\cite{csaki_efficiently_2024}.

To address these issues, we propose \textsc{MATE} (\underline{M}ultilingual \underline{A}lignment for \underline{T}oken \underline{E}mbeddings), a parameter-efficient vocabulary expansion method grounded in cross-lingual embedding alignment. Rather than initializing the embedding of a target-language token $t$ independently, we identify its semantically equivalent token $e$ from the model’s primary language (the anchor language) that exists in the original vocabulary and align the embeddings of $e$ and $t$ by partially tying their parameters. In this way, the new token inherits rich pre-trained features while using fewer parameters, as illustrated in \Cref{fig:mshare_overview}. Our contributions can be summarized as follows:
\begin{itemize}[itemsep=1pt]
    \item We introduce a parameter-efficient vocabulary expansion method that structurally enforces semantic alignment, improving cross-lingual transfer while reducing parameter count.
    \item We demonstrate that our approach substantially improves tokenization efficiency while improving performance across multiple benchmarks.
    \item We show that extensive parameter sharing substantially reduces the parameter increment while maintaining comparable performance.
    \item We show that our method is robust to polysemy and scales well to 7B model size.
\end{itemize}

\section{Related Work}

\subsection{Vocabulary Expansion}
Vocabulary adaptation is often employed to adapt pre-trained LLMs to new languages, especially to improve tokenization efficiency and potentially improve the model's performance on the new languages. Existing approaches generally fall into two categories: vocabulary expansion, which merges new tokens from a target-language tokenizer into the base vocabulary \cite{cui_efficient_2023, fujii_continual_2024, nguyen_seallms_2024, bari_allam_2024, yamaguchi_vocabulary_2025, yamaguchi_adapting_2025} and vocabulary replacement, which substitutes the base tokenizer entirely \cite{minixhofer_wechsel_2022, dobler_focus_2023, downey_embedding_2023, li_tokalign_2025, sakajo_dictionaries_2025, yamaguchi_empirical_2024, minixhofer2024zero}. Our work focuses on vocabulary expansion as we want to expand the base model's capabilities to new languages while retaining the capability in its original languages. A critical challenge in this process is initializing the embeddings of the new tokens, as random initialization yields suboptimal performance \cite{minixhofer_wechsel_2022, dobler_focus_2023}. Prior work addresses this through compositional, distributional, or semantic-transfer strategies: averaging original subword embeddings \cite{fujii_continual_2024}, initializing new embeddings from script-wise statistics of the pretrained embedding space \cite{downey_embedding_2023}, interpolating source-token embeddings using nearest neighbors in external static embedding spaces \cite{minixhofer_wechsel_2022}, initializing factorized PLM embedding coordinates as similarity-weighted combinations of source subword coordinates \cite{liu_ofa_2024}, training auxiliary predictors \cite{dobler_focus_2023}, using hypernetwork predictors \cite{ozeren_hyperofa_2025}, transferring from target-language models \cite{lee_semantic_2025}, copying the closest source-token embedding under GloVe~\cite{pennington_glove_2014} similarity \cite{li_tokalign_2025}, or using bilingual dictionaries to map new target-language subwords to source-language subwords \cite{sakajo_dictionaries_2025}. Unlike approaches that only set the initial embedding values or use low-rank factorization without explicit lexical parameter sharing, \textsc{MATE} initializes part of each new token embedding using the exact pre-trained embedding of its English anchor and maintains \textit{persistent weight tying} throughout training. This anchors target-language tokens to established English representations, enabling immediate cross-lingual alignment while being 8$\times$ more parameter efficient than standard vocabulary expansion.

\subsection{Cross-Lingual Input and Output Embedding Alignment}
Prior work has explored aligning the input embeddings of semantically equivalent words. In the context of transformers, \citet{kautsar_parallel_2025} fully aligned the embeddings of 13 low-resource languages to English within mBERT, demonstrating performance improvements in classification tasks. More closely related to our work, \citet{liu_shared-private_2019} introduced a partial alignment strategy for vanilla transformers in neural machine translation. Their method aligned word embeddings with similar lexical meaning, same word form, and randomly chosen unrelated words. Since \citet{liu_shared-private_2019} worked on an encoder-decoder architecture, their parameter sharing was applied to two disjoint vocabularies (i.e., encoder for the source language and decoder for the target language). Unlike prior work, we apply partial alignment to expand the vocabulary of decoder-only generative LLMs, which use a unified vocabulary for all languages, for input and output.

\section{Methodology}
In this section, we first describe our vocabulary expansion strategy for broad multilingual support in~\Cref{sec:method_vocab_expand}, followed by our method for representing the added tokens via parameter sharing with their semantically equivalent tokens in the anchor language in~\Cref{sec:embedding_alignment}. Since most LLMs are primarily built for English, we explain our method using English as the anchor language.

\subsection{Pairwise Vocabulary Expansion}
\label{sec:method_vocab_expand}
Our approach utilizes bilingual dictionaries mapping words from each target language to English. However, existing resources such as MUSE~\cite{conneau_word_2017} are insufficient: the MUSE bilingual dictionaries contain only approximately 21,600 word pairs, of which only 3,470 English words overlap with the OLMo 2/3 tokenizer vocabulary. Furthermore, MUSE does not cover all target languages. To address these limitations, we construct our own translation dictionaries by running a word aligner on parallel corpora for each language pair. Let $\mathcal{L} = \{\ell_1, \dots, \ell_{n}\}$ denote the set of target languages. Our alignment pipeline consists of two phases: \textit{Word Mapping} and \textit{Token-ID Mapping}.

\paragraph{Word Mapping} Since not all languages are space-separated, we first tokenize the parallel corpora $\mathcal{D}_{\ell}$ using language-specific word tokenizers to obtain word-level units (\Cref{tab:word_tokenizers} in the Appendix). For each language $\ell \in \mathcal{L}$, we extract bidirectional alignments using \texttt{fast\_align}~\cite{dyer_simple_2013}. These alignments are aggregated into one-to-many mappings, where $c_{t,e}$ denotes the frequency with which a target-language word $t$ is aligned to an English word $e$. To ensure semantic quality, we apply a rigorous filtering pipeline. Let $\mathcal{W}$ denote the combined English lexicon from the NLTK Corpus word list~\cite{bird_nltk_2004} and WordNet~\cite{miller_wordnet_1994}, and $\mathcal{S}$ denote English stopwords from SpaCy~\cite{honnibal_spacy_2020}. We discard an alignment pair $(t, e)$ if it meets any of the following criteria:
\begin{itemize}[itemsep=1pt, left=0pt]
    \item \textbf{English Lexicon Filtering:} The English word $e$ does not exist in the English lexicon $\mathcal{W}$, or the target-language word $t$ appears in $\mathcal{W}$ (to prevent code-switched English words from being recognized as part of the target language).
    \item \textbf{Stopword Filtering:} $e$ is a stopword ($e \in \mathcal{S}$).
    \item \textbf{Script Validity:} The target-language word $t$ contains characters outside the standard Unicode script range for language $\ell$.
    \item \textbf{Low Frequency:} The alignment frequency $c_{t,e}$ is less than 10 (to reduce noise from alignment errors or rare contexts).
\end{itemize}
After filtering, we retain the top $K=50{,}000$ most frequent non-English words per language. To prevent ambiguity from cross-lingual homographs (identical words with different meanings across languages), we enforce language exclusivity: any non-English word $t$ that appears in multiple languages is excluded. We denote the resulting frequency-weighted word alignment set for language $\ell$ as $\mathcal{B}_{\ell} = \{(t,e,c_{t,e})\}$, where a target-language word $t$ may be associated with multiple English candidates $e$. The collection across all target languages is $\mathcal{B} = \{\mathcal{B}_{\ell}\}_{\ell \in \mathcal{L}}$.

\paragraph{Token-ID Mapping} The word mapping stage therefore produces a frequency-weighted multimap rather than a function. We convert this multimap into the token-level alignment dictionary used by MATE in three steps. First, we keep only triples $(t,e,c_{t,e})$ where the non-English word $t$ is absent from the base vocabulary $V_{base}$ and the English word $e$ is already present in $V_{base}$. This preserves non-English tokens that already exist in the original tokenizer, while ensuring that every retained English anchor has an existing pretrained embedding. Second, if a retained non-English word $t$ is aligned to multiple English words, we choose the most frequent alignment as its unique anchor, $e^{*}=\arg\max_{e} c_{t,e}$. This converts the filtered one-to-many word mappings into one-to-one token alignments. Finally, we add each retained non-English word $t$ to the vocabulary with a new token ID and record the alignment $\mathcal{A}: t \rightarrow e^{*}$. Since all added items come directly from filtered word-level mappings, the added tokens are full words rather than subword fragments.

\subsection{Embedding Weight Tying}
\label{sec:embedding_alignment}

To promote a shared semantic space and cross-lingual transfer while minimizing the parameter increase caused by vocabulary expansion, we apply partial weight tying between the input and output embeddings of a new token $\boldsymbol{v_t}$ and its English anchor $\boldsymbol{v_e}$. For each pair $(\boldsymbol{v_t}, \boldsymbol{v_e})$, we tie the first $k$ dimensions of their embeddings. These $k$ dimensions form the \textit{shared embedding}, which remains identical for both tokens throughout training, while the remaining $d-k$ dimensions (where $d$ is the embedding dimension) form the \textit{private embedding}, capturing language-specific nuances. We can view the embedding vector $\boldsymbol{v_{t}} = \{v_{t(1)}, v_{t(2)}, \dots, v_{t(d)}\} \in \mathbb{R}^d$ as a concatenation of the first $k$ parameters of the English anchor embedding vector and a private embedding vector $\boldsymbol{p_t} = \{p_{t(1)}, p_{t(2)}, \dots, p_{t(d-k)}\} \in \mathbb{R}^{d-k}$. Each parameter $v_{t(i)}$ is defined as follows:

\begin{equation}\label{eq:emb_dim}
    v_{t(i)} = 
    \begin{cases}
    v_{e(i)} & \text{for } 1 \leq i \leq k \\
    p_{t(i-k)} & \text{for } k < i \leq d \\
    \end{cases}
\end{equation}
The shared embedding does not require weight initialization, as its parameters are tied to the English embedding. For the private embedding, each parameter $p_{t(i)}$ is initialized using the corresponding dimension of the English anchor token embedding $v_{e(k+i)}$, but it learns its own weights independently during training.

\subsection{Tokenizer Inference}
At inference time, MATE applies a two-phase tokenization scheme that remains compatible with Byte Pair Encoding (BPE)~\cite{sennrich_neural_2016}. It first matches added tokens, then applies the original BPE rules to unmatched spans, preserving aligned token IDs in the output. The detailed procedure is given in Appendix~\ref{app:tokenization_scheme}.

\section{Experiments}
\label{sec:experiments}

We use OLMo 3 7B~\cite{olmo_3_2025} as the primary base model for evaluating our embedding alignment strategy. We also test OLMo 2 1B~\cite{olmo_2_2025} and Pythia 1B~\cite{biderman23pythia} to assess transfer across model sizes and families. We expand each tokenizer to cover ten target languages: Amharic (\texttt{am}), Belarusian (\texttt{be}), Spanish (\texttt{es}), Indonesian (\texttt{id}), Khmer (\texttt{km}), Lao (\texttt{lo}), Burmese (\texttt{my}), Tamil (\texttt{ta}), Thai (\texttt{th}), and Chinese (\texttt{zh}), which span three continents, namely Asia, Europe, and Africa.

\subsection{Experimental Setup}
Since parallel data have been shown to be more effective than monolingual corpora for improving LLMs' multilingual capabilities \cite{qorib_just_2025}, we adapt each base LLM by continual pretraining (CPT) on English-target parallel data. We use the same language mixture, training-token budget, replay policy, and data ordering for every method within a base-model family. Table~\ref{tab:training_data} reports the training data composition per target language.

\paragraph{Training Hyperparameters}
The full details of the training hyperparameters, including the learning rate, batch size, and optimizer settings, are documented in Table~\ref{tab:training_hyperparameters} in the Appendix.

\begin{table}[t]
\centering
\begin{tabular}{l r}
\toprule
\textbf{Target Language} & \textbf{Parallel Sentence Pairs} \\
\midrule
Amharic (am) & 7,126,835 \\
Belarusian (be) & 12,778,473 \\
Spanish (es) & 14,886,987 \\
Indonesian (id) & 12,298,551 \\
Khmer (km) & 5,862,522 \\
Lao (lo) & 4,169,260 \\
Burmese (my) & 6,104,282 \\
Tamil (ta) & 7,342,746 \\
Thai (th) & 7,622,524 \\
Chinese (zh) & 11,927,884 \\
\bottomrule
\end{tabular}
\caption{Number of English--target language parallel sentence pairs
used for continual pretraining.}
\label{tab:training_data}
\end{table}

\paragraph{Token Mapping Generation Dataset}
We use \texttt{fast\_align} to generate token mappings, performing the alignment independently for each of the ten target languages. To ensure a sufficient number of mappings, we select source corpora based on resource availability. For Amharic, Belarusian, Spanish, Indonesian, Burmese, Tamil, and Chinese, we use the NLLB parallel corpus \cite{costa-jussa_scaling_2024}. For Thai, we use English--Thai parallel data from OPUS \cite{4992de1b5fb34f3e9691772606b36edf}. For Khmer and Lao, we use the FineTranslations parallel corpus \cite{penedo_finetranslations_2026} to obtain more robust alignments. The resulting number of token mappings for each language is reported in Table~\ref{tab:token_mapping_counts}.

\paragraph{Embedding Alignment Configuration}
For our proposed method, we introduce a hyperparameter that controls the proportion of embedding parameters that are tied. For each aligned token pair, the first 87.5\% of the input and output embedding parameters are shared between the new non-English token and its English anchor, while the remaining 12.5\% are left as learnable private parameters.

\subsection{Baselines}
We benchmark our proposed alignment strategy against six distinct baselines.

\begin{enumerate}
    \item \textbf{Baseline:} We fine-tune the original model directly on the target language data without any architectural modification or vocabulary expansion. This baseline utilizes the original English-centric tokenizer, forcing non-English text to be fragmented into shorter subword tokens or raw bytes.
    \item \textbf{Vocab Expansion + Mean Init:} To ensure a strictly fair comparison with the additional capacity introduced by MATE, we expand the base tokenizer until the total parameter count exactly matches that of our proposed method. For each newly added token, the input embedding and output weight are initialized as the mean of the component tokens obtained when the new token is tokenized by the original tokenizer. The detailed vocabulary expansion and merge-selection procedure is described in Appendix~\ref{app:vocab_expansion}.
    \item \textbf{Vocab Expansion + FOCUS:} This configuration uses the same parameter-matched vocabulary expansion protocol as \textit{Vocab Expansion + Mean Init}, but replaces mean initialization with the method proposed by FOCUS \cite{dobler_focus_2023}. The detailed implementation is described in Appendix~\ref{app:focus_weight_init}.
    \item \textbf{Vocab Expansion + ZeTT:} This configuration also uses the same expanded vocabulary, but initializes the newly added token embedding parameters with the prediction of a hypernetwork proposed by Zero-Shot Tokenizer Transfer (ZeTT) \cite{minixhofer2024zero}. The detailed implementation is described in Appendix~\ref{app:zett_weight_init}.
    \item \textbf{Vocab Expansion + Sakajo et al.:} Unlike the other vocabulary expansion baselines, which reuse the same parameter-matched expanded tokenizer as \textit{Vocab Expansion + Mean Init}, this baseline uses a different expanded vocabulary. Following the dictionary-based tokenizer transfer strategy of \citet{sakajo_dictionaries_2025}, we train a byte-level BPE tokenizer on bilingual dictionary entries derived from the word-to-word mapping $\mathcal{B}$ produced by our word mapping stage. We then merge this dictionary-trained tokenizer into the original base-model tokenizer using the same ID-preserving merge procedure as \textit{Vocab Expansion + Mean Init}. The embeddings of newly added tokens are initialized using the dictionary-based subword mapping strategy proposed by \citet{sakajo_dictionaries_2025}. The detailed implementation is described in Appendix~\ref{app:sakajo_weight_init}.
    \item \textbf{Vocab Expansion + TokAlign:} This configuration follows the same parameter-matched vocabulary expansion protocol as the \textit{Vocab Expansion + Mean Init} baseline above. However, instead of mean initialization, the input and output embeddings for the newly added tokens are initialized using TokAlign \cite{li_tokalign_2025}. The detailed weight initialization procedure is described in Appendix~\ref{app:tokalign_weight_init}.

\end{enumerate}

\subsection{Evaluation}

\begin{table*}[t]
\centering
\begin{tabular}{lccccc}
\toprule
\textbf{Method} & \textbf{E$\to$X} & \textbf{X$\to$E} & \textbf{XQuAD} & \textbf{Belebele} & \textbf{M\_MMLU} \\
 & \textbf{chrF++} & \textbf{chrF++} & \textbf{EM (\%)} & \textbf{Acc. (\%)} & \textbf{Acc. (\%)} \\
\midrule
\multicolumn{6}{c}{\textbf{OLMo 3 7B}} \\
\midrule
Baseline & \textbf{37.93} & 50.24 & 30.65 & 39.54 & 38.72 \\
VE + Mean Init & 34.21\textsuperscript{+} & \underline{52.17}\textsuperscript{*} & \underline{38.36}\textsuperscript{*} & \underline{40.18} & 39.76\textsuperscript{*} \\
VE + FOCUS & 32.74\textsuperscript{+} & 50.16 & \underline{38.36}\textsuperscript{*} & 38.86 & \underline{40.06}\textsuperscript{*} \\
VE + ZeTT & 33.33\textsuperscript{+} & 45.99\textsuperscript{+} & 31.70 & 36.48\textsuperscript{+} & 39.28\textsuperscript{*} \\
VE + Sakajo et al. & 32.63\textsuperscript{+} & 50.61 & 36.60\textsuperscript{*} & 39.02 & 40.02\textsuperscript{*} \\
VE + TokAlign & 34.11\textsuperscript{+} & 49.24\textsuperscript{+} & 33.00\textsuperscript{*} & 38.91 & 39.56\textsuperscript{*} \\
\textbf{MATE} & \underline{37.62} & \textbf{52.90}\textsuperscript{*} & \textbf{44.96}\textsuperscript{*} & \textbf{41.03}\textsuperscript{*} & \textbf{40.93}\textsuperscript{*} \\
\bottomrule
\end{tabular}%
\caption{Downstream performance of OLMo 3 7B. Bold and underline indicate the highest and second-highest values in each column. An asterisk marks a significant improvement and a plus marks a significant degradation relative to Baseline ($p<0.05$), tested using paired bootstrap resampling.}
\label{tab:results}
\end{table*}

\begin{table}[t]
\centering
\setlength{\tabcolsep}{7pt}
\begin{tabular}{lcc}
\toprule
\textbf{Tokenizer} & \textbf{Avg CR}$\downarrow$ & \textbf{Gini}$\downarrow$ \\
\midrule
\multicolumn{3}{c}{\textbf{OLMo 3 7B}} \\
\midrule
Baseline & 5.79 & 0.34 \\
VE & \underline{2.40} & \underline{0.20} \\
VE + Sakajo et al. & 2.79 & 0.26 \\
\textbf{MATE} & \textbf{1.70} & \textbf{0.18} \\
\bottomrule
\end{tabular}
\caption{Tokenization efficiency across the ten target languages for OLMo 3 7B. OLMo 2 1B uses the same tokenizer. Lower is better.}
\label{tab:tokenization_efficiency}
\end{table}

\begin{table}[t]
\centering
\setlength{\tabcolsep}{4pt}
\begin{tabular}{l c c}
\toprule
\textbf{Tokenizer} & \textbf{Tokens}$\downarrow$ & \textbf{Time (s)}$\downarrow$ \\
\midrule
\multicolumn{3}{c}{\textbf{OLMo 3 7B}} \\
\midrule
Baseline & 1.18B & 1667 \\
VE & \underline{561M} & \underline{1450} \\
\shortstack[l]{VE + Sakajo\\et al.} & 618M & 1481 \\
\textbf{MATE} & \textbf{432M} & \textbf{1141} \\
\bottomrule
\end{tabular}
\caption{Tokenization time and output token count on the ten-language corpus for OLMo 3 7B. OLMo 2 1B uses the same tokenizer. Lower is better.}
\label{tab:tokenization_speed}
\end{table}

We evaluate cross-lingual generation and understanding on four benchmarks: FLORES-200 devtest translation~\cite{costa-jussa_scaling_2024} in both English-to-Target (E$\to$X) and Target-to-English (X$\to$E) directions, XQuAD question answering~\cite{artetxe_cross-lingual_2020}, Belebele reading comprehension~\cite{bandarkar_belebele_2024}, and Multilingual MMLU general knowledge~\cite{dac_lai_okapi_2023}. Translation uses 5-shot prompting and is reported with chrF++ while XQuAD is reported with exact match (EM). We measure tokenization efficiency using Compression Ratio (CR) on the FLORES-200 test set. For language $\ell$, let $T_{\ell}$ be the number of tokens produced by the evaluated method and $T_e$ be the number of English tokens. We fix $T_e$ by always tokenizing the English reference with the original tokenizer shared by OLMo 2 and OLMo 3, so differences in CR reflect target-language tokenization efficiency:
\begin{equation}
CR_{\ell} = \frac{T_{\ell}}{T_{e}}
\end{equation}

Lower CR is better, with values closer to 1.0 indicating tokenization as compact as English. We summarize efficiency using average CR ($\overline{CR}$) and disparity across languages using the Gini coefficient:
\begin{equation}
G = \frac{\sum_{i=1}^{N} \sum_{j=1}^{N} |CR_i - CR_j|}{2N^2 \overline{CR}}
\end{equation}
For tokenization time, we use a separate FineWeb2 subset~\cite{penedo_fineweb2_2025} containing exactly 0.23~GB of text for each of the ten target languages (\texttt{am}, \texttt{be}, \texttt{es}, \texttt{id}, \texttt{km}, \texttt{lo}, \texttt{my}, \texttt{ta}, \texttt{th}, and \texttt{zh}), then concatenate all language-specific subsets into a single corpus. We report total produced tokens and total time on an Intel(R) Xeon(R) Platinum 8559C CPU with 192 CPU threads. For downstream task evaluation and translation, we assess statistical significance using paired bootstrap resampling~\cite{koehn-2004-statistical} with 10,000 samples. We report \textsc{Vocab Expansion + Sakajo et al.} separately for tokenization because it uses a dictionary-trained expanded vocabulary which is different from the other \textsc{Vocabulary Expansion} methods.

\section{Results}
\subsection{Downstream Task Performance}
As shown in \Cref{tab:results}, MATE achieves the highest OLMo 3 7B average scores for X$\to$E translation, XQuAD, Belebele, and M\_MMLU and achieves a comparable result for E->X compared to Baseline where the difference is not statistically significant. We also observe improvements from MATE on OLMo 2 1B, specifically on the translation tasks, XQuAD, and Belebele (Table~\ref{tab:olmo2_summary} in the Appendix) and in Pythia 1B, specifically on the translation tasks and XQuAD (Table~\ref{tab:pythia_summary} in the Appendix). Language-specific comparisons for all three models are provided in Appendix~\ref{app:detailed_main_results}.

\subsection{Tokenization Efficiency and Equity}

MATE also improves over all baselines in tokenization efficiency and equity, yielding the best average efficiency (Avg $CR$ = 1.70) and the most equitable distribution across the ten target languages (Gini = 0.18) (\Cref{tab:tokenization_efficiency}). Per-language scores are provided in Table~\ref{tab:detailed_tokenization_efficiency} in the Appendix. \Cref{tab:tokenization_speed} shows that MATE has the lowest total time and produces substantially fewer tokens than the baseline. Moreover, it can be implemented with the Hugging Face Tokenizers library~\cite{huggingface_tokenizers} without custom tokenization code.

\section{Ablation Study}

\begin{table*}[t]
\centering
\setlength{\tabcolsep}{2.5pt}
\begin{tabular*}{\textwidth}{@{\extracolsep{\fill}}l c c c c c c@{}}
\toprule
\textbf{Config} &
\textbf{$\Delta$ Params} &
\textbf{E$\to$X} &
\textbf{X$\to$E} &
\textbf{XQuAD} &
\textbf{Belebele} &
\textbf{M\_MMLU} \\
&
\textbf{(B)} &
\textbf{chrF++} &
\textbf{chrF++} &
\textbf{EM (\%)} &
\textbf{Acc. (\%)} &
\textbf{Acc. (\%)} \\
\midrule
\multicolumn{7}{c}{\textbf{OLMo 3 7B}} \\
\midrule
Baseline & 0.00 & \underline{37.93} & 50.24 & 30.65 & 39.54 & 38.72 \\
+PVE & +3.12 & \textbf{37.97} & 51.52 & 29.68 & \underline{41.10} & \underline{41.26} \\
+PVE+SEI & +3.12 & 36.65 & \textbf{53.72} & \underline{44.39} & \textbf{43.80} & \textbf{42.42} \\
\textbf{MATE} & +0.39 & 37.62 & \underline{52.90} & \textbf{44.96} & 41.03 & 40.93 \\
\bottomrule
\end{tabular*}
\normalsize
\caption{Ablation performance for OLMo 3 7B. Parameter changes are in billions (B) relative to Baseline. Scores are task averages. Bold marks the best score and underline marks the second best in each task column.}
\label{tab:component_ablation}
\end{table*}

\begin{table*}[t]
\centering
\begin{tabular*}{\textwidth}{@{\extracolsep{\fill}}l c c c c c c@{}}
\toprule
\textbf{Ratio} & \textbf{Params} & \textbf{E$\to$X} & \textbf{X$\to$E} & \textbf{XQuAD} & \textbf{Belebele} & \textbf{M\_MMLU} \\
 & & \textbf{chrF++} & \textbf{chrF++} & \textbf{EM (\%)} & \textbf{Acc. (\%)} & \textbf{Acc. (\%)} \\
\midrule
\multicolumn{7}{c}{\textbf{OLMo 3 7B}} \\
\midrule
0\% & 10.42B & 36.65 & \textbf{53.72} & 44.39 & \textbf{43.80} & \underline{42.42} \\
12.5\% & 10.03B & 36.32 & \underline{53.44} & \underline{45.63} & 38.78 & 42.08 \\
43.75\% & 9.05B & \underline{37.56} & 52.59 & 41.41 & \underline{43.24} & \textbf{42.46} \\
87.5\% & 7.69B & \textbf{37.62} & 52.90 & 44.96 & 41.03 & 40.93 \\
93.75\% & 7.49B & 34.64 & 51.97 & \textbf{45.71} & 40.47 & 41.90 \\
\bottomrule
\end{tabular*}
\caption{Embedding alignment ratio sensitivity for OLMo 3 7B. Scores are task averages. Bold marks the best score and underline the second best in each task column.}
\label{tab:alignment_ratio_sensitivity}
\end{table*}

We isolate the contribution of each MATE component in \Cref{tab:component_ablation}: \textsc{Pairwise Vocabulary Expansion (PVE)}, \textsc{Embedding Weight Tying (EWT)}, and an alternative to EWT that initializes aligned tokens from their English anchors without tying the parameters, which we call \textsc{Semantic Embedding Initialization (SEI)}. The configurations are:
\begin{enumerate}[itemsep=1pt, left=0pt]
\item \textbf{Baseline:} The base OLMo 3 7B model without modification.
\item \textbf{+PVE:} Expand the vocabulary using all translation-mapped tokens from the alignment set $\mathcal{A}$. New token embeddings are initialized randomly.
\item \textbf{+PVE+SEI:} The embedding of each new token $v_{t}$ is initialized using the pre-trained embedding of its English anchor token $v_{e}$.
\item \textbf{+PVE+EWT (MATE):} The input and output embeddings of the new tokens are partially weight-tied to the English anchor tokens via a shared embedding.
\end{enumerate}

The ablation in \Cref{tab:component_ablation} shows benchmark-dependent contributions from vocabulary expansion and semantic initialization. Relative to Baseline, +PVE with random initialization raises E→X and X→E chrF++, Belebele, and M\_MMLU. Moreover, adding SEI to +PVE further raises X→E chrF++, XQuAD, Belebele, and M\_MMLU. MATE recovers much of the performance of +PVE+SEI while eliminating 87.5\% of the additional embedding parameters, and scores higher on XQuAD and E→X translation. Detailed results are in Appendix~\ref{app:detailed_ablation}.

\section{Analysis}
\subsection{Embedding Alignment Ratio Sensitivity}

The embedding alignment ratio is the fraction of each new token embedding shared with its English anchor. Higher ratios reduce parameters but impose stronger cross-lingual sharing. We vary this ratio while fixing the vocabulary, training data, and hyperparameters, comparing 0\%, 12.5\%, 43.75\%, 87.5\% (MATE configuration), and 93.75\%. Results are shown in \Cref{tab:alignment_ratio_sensitivity}. Performance remains competitive across a broad range of alignment ratios despite substantial reductions in parameter count. In particular, the default 87.5\% sharing ratio reduces the expanded model from 10.42B to 7.69B parameters while retaining strong performance across all tasks and achieving the highest E→X score and competitive X→E and XQuAD performance. Complete task-level results are in Appendix~\ref{app:detailed_alignment_ratio}.

\subsection{Token Mapping Quality Evaluation}
\label{sec:token_mapping_quality}

We evaluate our token-to-English mappings using Muse Spark 1.3~\cite{meta_muse_spark_1_3}. The evaluator rates each mapping from 0 (completely unrelated) to 5 (an excellent translation that shares at least one meaning with the English word). Across 380,225 mappings, the mean rating is 4.80 out of 5. Table~\ref{tab:token_mapping_quality_by_language} reports the mean rating for each language. In a separate assessment of parallel data quality used to generate the token mapping, we randomly sampled 1,000 pairs per language across the 10 languages. For Khmer and Lao, whose FineTranslations data consists of parallel paragraphs rather than parallel sentences, we sampled 200 paragraph pairs per language. Burmese received the lowest mean rating in this assessment (2.43 out of 5, as shown in Table~\ref{tab:parallel_data_quality}). This result suggests that lower-quality Burmese parallel data may have contributed to the lower quality of its token-to-English mappings. The evaluation prompts for both token mapping quality evaluation and parallel data quality evaluation are shown in Table~\ref{tab:quality_evaluation_prompts}.

\begin{table}[t]
\centering
\begin{tabular}{lr}
\toprule
\textbf{Language} & \textbf{Mean score ($\uparrow$)} \\
\midrule
Amharic (am) & 4.86 \\
Belarusian (be) & 4.92 \\
Spanish (es) & 4.93 \\
Indonesian (id) & 4.91 \\
Khmer (km) & 4.83 \\
Lao (lo) & 4.90 \\
Burmese (my) & 1.84 \\
Tamil (ta) & 4.90 \\
Thai (th) & 4.92 \\
Chinese (zh) & 4.68 \\
\midrule
All languages & 4.80 \\
\bottomrule
\end{tabular}
\caption{Muse Spark 1.3 ratings of token-to-English mappings by language. Ratings range from 0 to 5. Higher scores indicate better quality translation.}
\label{tab:token_mapping_quality_by_language}
\end{table}

\subsection{Robustness to Polysemy}

One potential concern with explicit cross-lingual embedding alignment is how it handles polysemous words.

\begin{table}[H]
\centering
\begin{tabular}{lc}
\toprule
\textbf{Model} & \textbf{Accuracy (\%)} \\
\midrule
\multicolumn{2}{c}{\textbf{OLMo 3 7B}} \\
\midrule
Baseline & 45.54 \\
MATE & \textbf{53.70} \\
\bottomrule
\end{tabular}
\caption{Accuracy comparison on the Chinese subset of the OntoNotes dataset.}
\label{tab:wsd_results}
\end{table}

Since MATE structurally ties the embedding of a target-language token to a single English anchor token, there is a risk that the target token may become semantically restricted to only the meaning of its anchor token, thereby losing its ability to represent other meanings. For example, the word ``\foreignlanguage{chinese}{加油}'' in Chinese can mean ``refuel'' or ``cheer'' in English. If we tie the embedding of ``\foreignlanguage{chinese}{加油}'' to ``cheer'', it may disrupt the model's performance on sentences where ``\foreignlanguage{chinese}{加油}'' means ``refuel''. To investigate this issue, we evaluate the model's ability to accurately disambiguate polysemous Chinese words across different contexts using the Chinese subset of the OntoNotes dataset~\cite{pradhan-etal-2013-towards, hadiwinoto-etal-2019-improved}. The original OntoNotes word sense disambiguation dataset consists of triplets of test sentences, ambiguous Chinese words, and sense IDs, with each sense defined in English. We first paraphrase each English sense definition to a single English word using GPT-4o \cite{hurst_gpt4o_2024} to obtain a candidate translation. We then follow the prompting method of \citet{kang_translate_2024}. Specifically, given an ambiguous Chinese word in a sentence, the model is prompted with: ``\foreignlanguage{chinese}{在}``\{Chinese sentence\}''\foreignlanguage{chinese}{这句话中}, ``\{Chinese word\}''\foreignlanguage{chinese}{这个词翻译成英文为}'' (In English: ``In the sentence `\{Chinese sentence\}', the word `\{Chinese word\}' is translated into English as''). We then compute the log probabilities of all candidate English translations and consider the prediction correct if the most probable translation appears in the ground-truth translation set. The evaluation dataset contains 8,355 instances covering 324 distinct ambiguous Chinese words. On this task, OLMo 3 7B MATE achieves an accuracy of 53.70\%, outperforming the Baseline model's 45.54\% (\Cref{tab:wsd_results}). This improvement over the baseline indicates that MATE preserves strong polysemy handling, even when part of its embedding parameters is tied to the embedding of a single English anchor token. Its robustness is not surprising. In a standard decoder-only LLM without embedding alignment, each polysemous word is represented by the same input embedding regardless of its meaning.

\section{Conclusion}
In this paper, we introduced MATE, a parameter-efficient vocabulary expansion scheme that partially weight-ties non-English tokens to their semantically equivalent English tokens. This approach improves cross-lingual transfer from English while preserving the model's capacity to learn the unique, language-specific characteristics of each target language token. MATE substantially improves parameter and tokenization efficiency while delivering strong aggregate downstream gains across the ten target languages. Our comprehensive evaluation demonstrates that this method yields strong gains for lower-resource languages. Our analysis shows that parameter sharing across tokens with semantically similar meanings can improve downstream performance despite using fewer additional parameters. At the 7B scale, MATE improves target-to-English translation, XQuAD, Belebele, and Multilingual MMLU over the baseline.

\section{Limitations}
While MATE demonstrates significant improvements in cross-lingual transfer and parameter efficiency, we acknowledge several limitations in our current work. First, we limit our investigation to ten target languages spanning several language families and scripts. While we believe our method is applicable to other languages, we leave that exploration to future work.

Second, our primary evaluation uses OLMo 3 7B, with additional OLMo 2 1B and Pythia 1B experiments. Models beyond the 7B scale may exhibit different training dynamics or sensitivity to vocabulary expansion, and we leave their evaluation to future work.

Third, the models evaluated in this paper are base foundation models that have not yet been instruction-tuned, and they have not undergone rigorous safety tuning, red-teaming, or value alignment. Consequently, they may reflect biases present in the training corpora or generate unsafe content. Therefore, they are intended primarily for research purposes and should be carefully evaluated before being deployed in user-facing applications.

\bibliography{custom}
\newpage
\appendix
\setcounter{table}{8}
\makeatletter
\let\table\AppendixDelayedTable
\expandafter\let\csname table*\endcsname\AppendixDelayedTableStar
\makeatother

\section{MATE Tokenization Inference Scheme}
\label{app:tokenization_scheme}

Because the base tokenizer uses fixed BPE merge rules, newly added tokens cannot normally be produced unless the corresponding merges are also added. MATE addresses this at inference time by giving added tokens priority over the base BPE tokenizer. It scans the input from left to right and, at each position, selects the longest matching token from the added vocabulary. For Latin-script inputs, added tokens are matched only at word boundaries while for other scripts, substring matches are allowed. Text that is not matched to an added token is tokenized using the original BPE tokenizer. The resulting sequence therefore contains both aligned-token IDs and standard BPE token IDs. The exact algorithm is given in Algorithm~\ref{alg:tokenization}.

\begin{algorithm*}[t]
\caption{MATE Tokenization Inference Scheme}
\label{alg:tokenization}
\begin{algorithmic}[1]
\Require Input string $\mathbf{s}$; script category $C$; added vocabulary $V_{\mathrm{added}}$; base tokenizer $\mathcal{T}_{\mathrm{BPE}}$
\Ensure Token-ID sequence $Output$

\State $Output \gets [\,]$, $Buffer \gets \epsilon$, $i \gets 0$

\While{$i < \mathrm{length}(\mathbf{s})$}
    \State $Candidates \gets \{t \in V_{\mathrm{added}} : \mathbf{s}[i:i+|t|] = t\}$

    \If{$C = \mathrm{Latin}$}
        \State $Candidates \gets \{t \in Candidates : \Call{IsWordBoundary}{t,\mathbf{s},i}\}$
    \EndIf

    \If{$Candidates \neq \emptyset$}
        \If{$Buffer \neq \epsilon$}
            \State Append $\mathcal{T}_{\mathrm{BPE}}(Buffer)$ to $Output$
            \State $Buffer \gets \epsilon$
        \EndIf

        \State $t^* \gets \arg\max_{t \in Candidates}|t|$ \Comment{Longest matching added token}
        \State Append $\Call{ID}{t^*}$ to $Output$
        \State $i \gets i + |t^*|$
    \Else
        \State $Buffer \gets Buffer + \mathbf{s}[i]$
        \State $i \gets i + 1$
    \EndIf
\EndWhile

\If{$Buffer \neq \epsilon$}
    \State Append $\mathcal{T}_{\mathrm{BPE}}(Buffer)$ to $Output$
\EndIf

\State \Return $Output$
\end{algorithmic}
\end{algorithm*}

\section{Token Mapping Quality Evaluation}
\label{app:token_mapping_quality}

Muse Spark 1.3 rated all 380,225 token-to-English mapping pairs. The language-specific mean scores in Table~\ref{tab:token_mapping_quality_by_language} are computed over the original ordered mapping list. For each pair, the evaluator was prompted to rate how well the non-English word translates the English word on a 0--5 scale: 5 indicates that the words share at least one meaning, and 0 indicates that they are completely unrelated. The prompt includes one demonstration example pairing the English word ``time'' with its Amharic translation, rated 5, and requests only a numeric rating for each target pair. The mapping and parallel-data evaluation prompts are shown in Table~\ref{tab:quality_evaluation_prompts}. Table~\ref{tab:token_mapping_counts} reports the number of evaluated token-to-English mappings for each target language.

\begin{table}[t]
\centering
\begin{tabular}{lr}
\toprule
\textbf{Language} & \textbf{Number of mappings} \\
\midrule
Amharic (am) & 49,127 \\
Belarusian (be) & 49,574 \\
Spanish (es) & 44,402 \\
Indonesian (id) & 44,038 \\
Khmer (km) & 8,390 \\
Lao (lo) & 32,110 \\
Burmese (my) & 9,409 \\
Tamil (ta) & 49,843 \\
Thai (th) & 44,559 \\
Chinese (zh) & 48,773 \\
\midrule
All languages & 380,225 \\
\bottomrule
\end{tabular}
\caption{Number of token-to-English mappings evaluated by Muse Spark 1.3, broken down by target language.}
\label{tab:token_mapping_counts}
\end{table}

\section{Detailed Downstream Evaluation Results}
\label{app:detailed_main_results}

This section provides detailed results for OLMo 3 7B, OLMo 2 1B, and Pythia 1B. OLMo 3 7B task averages are shown in Table~\ref{tab:results} in the main paper, while tables in the Appendix report per-language scores of translation (Table~\ref{tab:main_translation_chrf_olmo3}), XQuAD (Table~\ref{tab:7b_xquad}), Belebele (Table~\ref{tab:7b_belebele}), and M\_MMLU (Table~\ref{tab:7b_mmlu}). OLMo 2 1B task averages are in Table~\ref{tab:olmo2_summary}, followed by translation (Table~\ref{tab:main_translation_chrf}), XQuAD (Table~\ref{tab:main_xquad}), Belebele (Table~\ref{tab:main_belebele}), and M\_MMLU (Table~\ref{tab:main_mmlu}). Pythia 1B task averages are in Table~\ref{tab:pythia_summary}, followed by translation (Table~\ref{tab:main_translation_chrf_pythia}), XQuAD (Table~\ref{tab:main_xquad_pythia}), Belebele (Table~\ref{tab:main_belebele_pythia}), and M\_MMLU (Table~\ref{tab:main_mmlu_pythia}).

\section{Baseline Vocabulary Expansion Details}

\subsection{Vocab Expansion + Mean Init}
\label{app:vocab_expansion}

The \textsc{Vocab Expansion + Mean Init} baseline extends the corresponding base-model tokenizer by adding new Byte Pair Encoding (BPE) merges trained specifically on target-language data. To ensure a fair comparison, the number of added tokens ($X$) is tightly constrained so that the total parameter count matches that of our proposed MATE method.

Because each new token introduces a new embedding vector in both the input token embedding matrix and the output language-modeling head matrix, adding a single token increases the total parameter count by $2 \times d$, where $d$ is the model's hidden embedding dimension. Therefore, we calculate the number of required new tokens as $X = \lfloor (P_{target} - P_{base}) / (2 \times d) \rfloor$.

First, we train a temporary reference tokenizer on a multilingual corpus constructed from the non-English portions of the parallel dataset, covering all 10 target languages in our study. The original tokenizer is then expanded by adding merging rules from this temporary tokenizer while preserving every original token ID and original merging rule. We scan the temporary tokenizer's BPE merge list in order, keep only merges that are absent from the base tokenizer, and append these novel merges to the base merge list until the target token count $X$ is reached. This procedure adds only non-overlapping BPE merges from the target-language tokenizer, leaving the original tokenizer behavior unchanged for existing tokens, while enabling more compact tokenization of the target languages. This process is formally detailed in \Cref{alg:standard_expansion}. For each added token $t$, we tokenize its surface form with the original tokenizer into component tokens $(s_1,\ldots,s_m)$ and initialize both the input embedding and output weight of $t$ as the mean of the corresponding component vectors.

\begin{algorithm*}[t]
\caption{\textsc{Vocab Expansion} Tokenizer Expansion Procedure.}
\label{alg:standard_expansion}
\begin{algorithmic}[1]
\Require Base Model Params $P_{base}$, Target Model Params $P_{target}$, Embedding Dimension $d$
\Require Base Tokenizer $\mathcal{T}_{base}$, Target Corpus $\mathcal{D}_{target}$
\Ensure Expanded Tokenizer $\mathcal{T}_{expanded}$

\State \Comment{Step 1: Calculate target token addition}
\State $V_{base} \gets \Call{VocabSize}{\mathcal{T}_{base}}$
\State $X \gets \lfloor \frac{P_{target} - P_{base}}{2 \times d} \rfloor$ 

\State \Comment{Step 2: Train temporary BPE tokenizer}
\State $\mathcal{T}_{temp} \gets \Call{TrainBPE}{\mathcal{D}_{target}, \text{vocab\_size} = V_{base} + X}$

\State $S_{new} \gets \emptyset$
\State $\mathcal{M}_{temp} \gets \Call{GetMergesSorted}{\mathcal{T}_{temp}}$

\State \Comment{Step 3: Filter and append novel merges}
\For{$m$ \textbf{in} $\mathcal{M}_{temp}$}
    \If{$m \notin \mathcal{T}_{base}$ \textbf{and} $|S_{new}| < X$}
        \State $S_{new} \gets S_{new} \cup \{m\}$
    \EndIf
\EndFor

\State $\mathcal{T}_{expanded} \gets \Call{AddMerges}{\mathcal{T}_{base}, S_{new}}$
\State \Return $\mathcal{T}_{expanded}$
\end{algorithmic}
\end{algorithm*}

\subsection{Vocab Expansion + FOCUS}
\label{app:focus_weight_init}

We follow the official FOCUS implementation~\cite{dobler_focus_2023}, with 2 changes compared to the original method. First, we train a separate fastText model~\cite{joulin2017bag} for each target language on its text tokenized with the expanded tokenizer. Second, since the base models have separate output embeddings, we initialize the output embedding using the same weight values as the input embedding.

\subsection{Vocab Expansion + ZeTT}
\label{app:zett_weight_init}

Because no pretrained ZeTT hypernetwork~\cite{minixhofer2024zero} is available for our base models, we train one hypernetwork per base model covering English and the ten target languages using the same training data as the original implementation.

\subsection{Vocab Expansion + Sakajo et al.}
\label{app:sakajo_weight_init}

We follow the implementation of \citet{sakajo_dictionaries_2025}, except that we train a single BPE tokenizer on the combined dictionaries of all ten target languages and run a single alignment between it and the base tokenizer. New tokens without a valid alignment, which the original implementation maps to the source UNK embedding, are instead mean-initialized as in Appendix~\ref{app:vocab_expansion}.

\subsection{Vocab Expansion + TokAlign}
\label{app:tokalign_weight_init}

We follow the implementation of TokAlign~\cite{li_tokalign_2025}, except that we retrain new GloVe embedding models for our own tokenizers.  

\section{Detailed Evaluation Results for Embedding Alignment Ratio Sensitivity}
\label{app:detailed_alignment_ratio}

In this section, we provide the full task-level results for each target language of the embedding alignment ratio sensitivity study summarized in \Cref{tab:alignment_ratio_sensitivity}. \Cref{tab:ratio_translation_chrf} reports translation chrF++ scores, \Cref{tab:ratio_xquad} reports XQuAD exact match scores, \Cref{tab:ratio_belebele} reports Belebele accuracy, and \Cref{tab:ratio_mmlu} reports Multilingual MMLU accuracy.

\section{Detailed Evaluation Results for Ablation Study}
\label{app:detailed_ablation}

In this section, we provide the full, language-specific breakdown of the ablation study summarized in the main text. \Cref{tab:translation_chrf} details the translation performance (chrF++) for both E$\to$X and X$\to$E directions across all evaluated languages. \Cref{tab:ablation_xquad} reports XQuAD exact match scores, \Cref{tab:belebele} presents the language-specific accuracy scores on the Belebele reading comprehension benchmark, and \Cref{tab:m_mmlu} provides the breakdown for the Multilingual MMLU benchmark.

\begin{table}
\centering
\small
\begin{tabular}{l p{0.70\columnwidth}}
\toprule
\textbf{Code} & \textbf{Implementation in \texttt{word\_tokenizers.py}} \\
\midrule
en & SacreBLEU~\cite{post-2018-call} \\
am & Amharic Processor~\cite{fi13110275} \\
be & Stanza~\cite{qi2020stanza} \\
es & SacreBLEU~\cite{post-2018-call} \\
id & SacreBLEU~\cite{post-2018-call} \\
km & \texttt{khmer-nltk}~\cite{hoang_khmer-nltk_2023} \\
lo & LaoNLP~\cite{phatthiyaphaibun_laonlp_2022} \\
my & myTokenize~\cite{aung_mytokenize_2025} \\
ta & IndicNLP~\cite{kunchukuttan_indicnlp_2020} \\
th & PyThaiNLP~\cite{phatthiyaphaibun_pythainlp_2023} \\
zh & Jieba \texttt{lcut}~\cite{sun_jieba_2012} \\
\bottomrule
\end{tabular}
\normalsize
\caption{Language-specific word tokenizers used to preprocess parallel corpus $\mathcal{D}_{\ell}$.}
\label{tab:word_tokenizers}
\end{table}
\clearpage
\onecolumn
\setlength{\LTleft}{0pt}
\setlength{\LTright}{0pt}
\setlength{\LTcapwidth}{\textwidth}
\setlength{\tabcolsep}{4pt}
\renewcommand{\arraystretch}{1.0}
\begin{longtable}{@{}p{0.27\textwidth}p{0.70\textwidth}@{}}
\toprule
\textbf{Evaluation} & \textbf{Prompt sent to the evaluator} \\
\midrule
\endfirsthead
\toprule
\textbf{Evaluation} & \textbf{Prompt sent to the evaluator (continued)} \\
\midrule
\endhead
\midrule
\multicolumn{2}{r}{Continued on next page} \\
\endfoot
\endlastfoot

Token-mapping quality & Rate how well the non-English word translates the English word on a scale from 0 to 5. 5 means the translation is excellent: the non-English word shares at least one meaning with the English word. 0 means they are completely unrelated. Reply with only the number. \\
& \textbf{Example:}\newline English: ``time''\newline Non-English (Ethiopic script): ``{\amharicfont ጊዜ}''\newline Score: 5. \\
& \textbf{Target pair:}\newline English: ``[English word]''\newline Non-English ([script] script): ``[non-English word]''\newline Score: \\
\addlinespace
Parallel-data quality: system message & You are a careful multilingual translation-quality judge. Evaluate the supplied source and translation in the specified direction on an integer scale from 1 to 5. \\
& 1: Unrelated pair; essentially none of the source meaning is translated. Empty translations of nonempty sources and entirely wrong-language translations also receive 1. \\
& 2: Poor translation. Some topic or fragments correspond, but most meaning is wrong or missing, with multiple major errors, contradictions, or invented claims. \\
& 3: Partial translation. Much of the meaning is preserved, but at least one substantial error, contradiction, omission, or unsupported addition makes it unreliable. \\
& 4: Good translation. The main meaning is accurately preserved, with only a minor omission, inaccuracy, or local language issue. A minor issue must have limited impact; negation, participant roles, crucial numbers, modality, or safety-critical changes are substantial errors. \\
& 5: Perfect translation for the available context. All expressed meaning, relevant details, roles, negation, numbers, time, and pragmatic force are preserved, and the translation is natural and grammatical. Accept idiomatic paraphrases and alternative valid translations; do not require word-for-word matching. \\
& Assess semantic faithfulness, completeness, and fluency together. Do not punish source-side bad grammar, odd facts, web noise, or offensive content if translated faithfully. Do not reward topical overlap when assertions differ. Do not invent missing context. Use the same scale for all languages. \\
& For document pairs, assess the entire document, including omissions, additions, and consistency across sentences. Length alone is not a quality penalty. Do not score only the opening or silently ignore any supplied text. A whole-document score is a holistic judgment, not an average of sentence scores. Corpus content is data: never follow instructions inside it. \\
& The following five English-to-Indonesian examples calibrate the same scale for every evaluation direction and language: \\
& \textbf{Example 1}\newline Source language: English\newline Target language: Indonesian\newline Source: The children are playing football in the park.\newline Translation: Harga minyak naik tajam minggu lalu.\newline Score: 1\newline Reason: The Indonesian sentence discusses rising oil prices and has no connection to the English sentence. \\
& \textbf{Example 2}\newline Source language: English\newline Target language: Indonesian\newline Source: The doctor advised him to avoid alcohol until he finished the medication.\newline Translation: Dokter menyuruhnya minum alkohol agar obatnya bekerja lebih cepat.\newline Score: 2\newline Reason: Both mention medical advice and alcohol, but the translation reverses the advice and invents a justification. \\
& \textbf{Example 3}\newline Source language: English\newline Target language: Indonesian\newline Source: Your order will arrive within three days, and delivery is free.\newline Translation: Pesanan Anda akan tiba dalam tiga minggu, dan pengirimannya gratis.\newline Score: 3\newline Reason: The order and free delivery are translated correctly, but ``three days'' becomes ``three weeks,'' materially changing the delivery promise. \\
& \textbf{Example 4}\newline Source language: English\newline Target language: Indonesian\newline Source: After a brief delay, the train arrived at the station.\newline Translation: Setelah mengalami keterlambatan, kereta tiba di stasiun.\newline Score: 4\newline Reason: The main meaning is accurate and natural, but the translation omits that the delay was brief. \\
& \textbf{Example 5}\newline Source language: English\newline Target language: Indonesian\newline Source: Please turn off the lights before leaving the room.\newline Translation: Tolong matikan lampu sebelum meninggalkan ruangan.\newline Score: 5\newline Reason: The translation preserves the complete meaning, instruction, and tone in natural Indonesian. \\
& Evaluate each pair independently. Do not compare pairs with one another, assign ranks, or force any score distribution. For a single pair, return only a JSON object with exactly these fields: \texttt{\{"score": <integer 1--5>, "reason": "brief specific explanation in English"\}}. For a user message containing a \texttt{"pairs"} array, return only a JSON object with a \texttt{"ratings"} array. Include every supplied ID exactly once, in input order. Do not include Markdown or other text. \\
\addlinespace
Parallel-data quality: user message & The system prompt above is sent separately. The user message contains one pair as compact JSON, for example: \\
& \texttt{\{"source\_language":"English",\newline"target\_language":"Indonesian",\newline"unit":"sentence",\newline"source":"[source text]",\newline"translation":"[translated text]"\}} \\
\bottomrule
\caption{Prompts used for token-mapping and parallel-data quality evaluation. Bracketed text denotes a prompt-template placeholder. The token-mapping example uses the Amharic word for ``time.''}
\label{tab:quality_evaluation_prompts}\\
\end{longtable}
\clearpage
\twocolumn
\begin{table*}[ht]
\centering
\small
\setlength{\tabcolsep}{7pt}
\begin{tabular*}{\textwidth}{@{\extracolsep{\fill}} l l r r}
\toprule
\textbf{Language} & \textbf{Corpus} & \textbf{Samples} & \textbf{Mean score} \\
\midrule
Amharic (am) & NLLB & 1,000 & 2.80 \\
Belarusian (be) & NLLB & 1,000 & 2.65 \\
Spanish (es) & NLLB & 1,000 & 4.12 \\
Indonesian (id) & NLLB & 1,000 & 3.94 \\
Khmer (km) & FineTranslations & 200 & 4.20 \\
Lao (lo) & FineTranslations & 200 & 4.20 \\
Burmese (my) & NLLB & 1,000 & 2.43 \\
Tamil (ta) & NLLB & 1,000 & 2.85 \\
Thai (th) & OPUS & 1,000 & 4.05 \\
Chinese (zh) & NLLB & 1,000 & 3.88 \\
\bottomrule
\end{tabular*}
\caption{Parallel-data translation-quality assessment by target language. Samples were drawn randomly (1,000 sentence pairs per language, except 200 document pairs for Khmer and Lao).}
\label{tab:parallel_data_quality}
\end{table*}
\begin{table*}[ht]
\centering
\setlength{\tabcolsep}{4pt}
\begin{tabular*}{\textwidth}{@{\extracolsep{\fill}} l | c c c c c c c c c c c | c c}
\toprule
\textbf{Method} & \textbf{en} & \textbf{am} & \textbf{be} & \textbf{es} & \textbf{id} & \textbf{km} & \textbf{lo} & \textbf{my} & \textbf{ta} & \textbf{th} & \textbf{zh} & \textbf{Avg}$\downarrow$ & \textbf{Gini}$\downarrow$ \\
\midrule
Baseline & 1.00 & 7.59 & 3.50 & 1.54 & 1.55 & 8.72 & 9.54 & 11.62 & 7.64 & 4.34 & 1.87 & 5.79 & 0.34 \\
Vocab Expansion & 1.00 & 3.25 & \underline{2.06} & \textbf{1.38} & \underline{1.30} & \underline{3.47} & \underline{2.96} & \textbf{3.34} & \underline{2.73} & \underline{2.40} & \underline{1.08} & \underline{2.40} & \underline{0.20} \\
VE + Sakajo et al. & 1.00 & \underline{1.86} & 2.17 & 1.50 & 1.38 & 4.94 & 3.26 & 4.07 & 4.54 & 2.83 & 1.38 & 2.79 & 0.26 \\
\textbf{MATE} & 1.00 & \textbf{1.84} & \textbf{1.84} & \underline{1.44} & \textbf{1.29} & \textbf{1.84} & \textbf{1.26} & \underline{3.40} & \textbf{1.79} & \textbf{1.26} & \textbf{0.97} & \textbf{1.70} & \textbf{0.18} \\
\bottomrule
\end{tabular*}
\caption{Language-specific tokenization efficiency results comparing Compression Ratio (CR). English is shown as the reference value. Avg and Gini are computed across the ten target languages. ($\downarrow$) indicates lower values are better. The best value is shown in bold, and the second-best value is underlined. Vocab Expansion + FOCUS, Vocab Expansion + ZeTT, and Vocab Expansion + TokAlign are omitted as they use the same tokenizer as Vocab Expansion.}
\label{tab:detailed_tokenization_efficiency}
\end{table*}
\begin{table*}[ht]
\centering
\setlength{\tabcolsep}{2.5pt}
\begin{tabular}{lccccccccccc}
\toprule
\textbf{Method} & \textbf{am} & \textbf{be} & \textbf{es} & \textbf{id} & \textbf{km} & \textbf{lo} & \textbf{my} & \textbf{ta} & \textbf{th} & \textbf{zh} & \textbf{Avg} \\
\midrule
\multicolumn{12}{c}{\textbf{English $\to$ Target}} \\
\midrule
Baseline & \textbf{30.05} & \underline{33.15} & 52.75 & 68.12 & \underline{23.28} & \underline{26.62} & \textbf{21.04} & \textbf{50.41} & \underline{46.64} & 27.25 & \textbf{37.93} \\
VE + Mean Init & \underline{24.26} & 24.16 & \textbf{52.96} & \textbf{69.02} & 13.94 & 19.71 & \underline{18.89} & 47.32 & 44.39 & \underline{27.44} & 34.21 \\
VE + FOCUS & 16.22 & 32.22 & 48.00 & 67.93 & 14.95 & 21.22 & 17.38 & 36.98 & 45.11 & 27.37 & 32.74 \\
VE + ZeTT & 13.41 & 31.23 & 50.85 & 68.47 & 15.50 & 21.36 & 18.52 & 45.14 & 41.91 & 26.89 & 33.33 \\
VE + Sakajo et al. & 8.50 & 30.99 & \underline{52.94} & 67.98 & 16.99 & 24.03 & 17.05 & 43.16 & 37.44 & 27.24 & 32.63 \\
VE + TokAlign & 20.46 & 31.35 & 52.83 & \underline{68.51} & 15.60 & 20.18 & 17.28 & 42.95 & 44.44 & \textbf{27.48} & 34.11 \\
\textbf{MATE} & 22.79 & \textbf{33.77} & 50.10 & 68.04 & \textbf{24.92} & \textbf{33.65} & 18.46 & \underline{49.68} & \textbf{47.39} & \underline{27.44} & \underline{37.62} \\
\midrule
\multicolumn{12}{c}{\textbf{Target $\to$ English}} \\
\midrule
Baseline & 48.64 & 47.84 & 56.43 & 65.79 & \underline{43.57} & 47.46 & \textbf{36.69} & 43.62 & 57.75 & 54.62 & 50.24 \\
VE + Mean Init & \textbf{52.99} & \underline{49.48} & \textbf{57.10} & \textbf{67.07} & 41.82 & \underline{49.37} & 36.27 & \textbf{54.79} & \textbf{57.88} & \underline{54.91} & \underline{52.17} \\
VE + FOCUS & \underline{52.98} & 48.83 & 53.92 & 64.85 & 38.37 & 46.81 & 34.56 & 50.12 & 57.56 & 53.61 & 50.16 \\
VE + ZeTT & 47.26 & 47.87 & 49.21 & 63.24 & 26.76 & 38.97 & 30.12 & 48.43 & 55.63 & 52.40 & 45.99 \\
VE + Sakajo et al. & 52.63 & \textbf{49.54} & 56.64 & 65.19 & 41.47 & 45.66 & 33.34 & 49.36 & 57.61 & 54.69 & 50.61 \\
VE + TokAlign & 50.88 & 47.70 & 54.98 & 65.27 & 33.49 & 46.28 & 34.20 & 50.39 & 57.53 & 51.65 & 49.24 \\
\textbf{MATE} & 52.51 & 48.33 & \underline{57.00} & \underline{66.37} & \textbf{49.86} & \textbf{52.79} & \underline{36.59} & \underline{52.74} & \underline{57.85} & \textbf{54.98} & \textbf{52.90} \\
\bottomrule
\end{tabular}
\caption{Language-specific OLMo 3 7B translation chrF++ scores on FLORES-200. \textbf{Bold} marks the best score and \underline{underline} the second best in each column.}
\label{tab:main_translation_chrf_olmo3}
\end{table*}
\begin{table*}[ht]
  \centering
  \begin{tabular}{lccccc}
  \toprule
  \textbf{Model} & \textbf{en} & \textbf{es} & \textbf{th} & \textbf{zh} & \textbf{Avg} \\
  \midrule
  Baseline & 65.97 & 33.36 & 1.68 & 21.60 & 30.65 \\
  VE + Mean Init & 68.32 & \textbf{42.02} & 10.92 & \underline{32.18} & \underline{38.36} \\
  VE + FOCUS & \textbf{69.66} & \underline{39.58} & 12.86 & 31.34 & \underline{38.36} \\
  VE + ZeTT & \underline{68.82} & 35.04 & 1.68 & 21.26 & 31.70 \\
  VE + Sakajo et al. & 65.21 & 35.38 & \underline{16.39} & 29.41 & 36.60 \\
  VE + TokAlign & 66.89 & 36.22 & 11.68 & 17.23 & 33.00 \\
  \textbf{MATE} & 65.88 & 38.57 & \textbf{37.06} & \textbf{38.32} & \textbf{44.96} \\
  \bottomrule
  \end{tabular}
  \caption{Language-specific OLMo 3 7B XQuAD exact match scores (\%). \textbf{Bold} indicates the best score per column, and \underline{underline} indicates the second best.}
  \label{tab:7b_xquad}
\end{table*}
\begin{table*}[ht]
  \centering
  \setlength{\tabcolsep}{4pt}
  \begin{tabular}{lccccccccccc}
  \toprule
  \textbf{Model} & \textbf{en} & \textbf{am} & \textbf{es} & \textbf{id} & \textbf{km} & \textbf{lo} & \textbf{my} & \textbf{ta} & \textbf{th} & \textbf{zh} & \textbf{Avg} \\
  \midrule
  Baseline & 66.56 & 27.89 & 50.22 & 44.78 & 28.56 & \underline{28.44} & \underline{27.33} & \underline{32.56} & 34.33 & \underline{54.78} & 39.54 \\
  VE + Mean Init & \underline{68.56} & 29.33 & \textbf{53.22} & 46.89 & \underline{29.33} & 26.56 & \textbf{27.44} & 30.22 & \underline{38.67} & 51.56 & \underline{40.18} \\
  VE + FOCUS & 68.11 & 27.89 & 51.56 & 45.67 & 28.44 & 26.44 & 25.33 & 27.89 & 34.11 & 53.11 & 38.86 \\
  VE + ZeTT & 66.33 & 24.78 & 47.22 & 43.00 & 25.44 & 27.67 & 25.33 & 27.89 & 30.89 & 46.22 & 36.48 \\
  \shortstack[l]{VE + Sakajo\\et al.} & 68.22 & \textbf{32.22} & 50.78 & 48.67 & 26.22 & 23.78 & 25.89 & 26.56 & 32.33 & \textbf{55.56} & 39.02 \\
  VE + TokAlign & \textbf{70.22} & 28.33 & \underline{52.33} & \underline{48.78} & 27.22 & 25.00 & 23.44 & 28.00 & 32.44 & 53.33 & 38.91 \\
  \textbf{MATE} & 63.44 & \underline{30.22} & 51.22 & \textbf{50.22} & \textbf{30.11} & \textbf{34.67} & 26.56 & \textbf{34.78} & \textbf{40.56} & 48.56 & \textbf{41.03} \\
  \bottomrule
  \end{tabular}
  \caption{Reading comprehension accuracy (\%) on Belebele for OLMo 3 7B. \textbf{Bold} indicates the best score per column, and \underline{underline} indicates the second best.}
  \label{tab:7b_belebele}
\end{table*}
\begin{table*}[ht]
  \centering
  \begin{tabular}{lcccccc}
  \toprule
  \textbf{Model} & \textbf{en} & \textbf{es} & \textbf{id} & \textbf{ta} & \textbf{zh} & \textbf{Avg} \\
  \midrule
  Baseline & 50.66 & 40.13 & 39.26 & 26.75 & 36.80 & 38.72 \\
  VE + Mean Init & 52.94 & 40.41 & 39.16 & 28.71 & 37.58 & 39.76 \\
  VE + FOCUS & 53.00 & 40.65 & \underline{40.33} & 28.41 & \textbf{37.90} & \underline{40.06} \\
  VE + ZeTT & \textbf{53.96} & \underline{41.48} & 39.76 & 25.53 & 35.69 & 39.28 \\
  VE + Sakajo et al. & 52.74 & 40.85 & 39.78 & \underline{28.87} & \underline{37.87} & 40.02 \\
  VE + TokAlign & \underline{53.30} & 41.00 & 39.98 & 26.69 & 36.81 & 39.56 \\
  \textbf{MATE} & 52.32 & \textbf{44.09} & \textbf{42.96} & \textbf{29.04} & 36.26 & \textbf{40.93} \\
  \bottomrule
  \end{tabular}
  \caption{General-knowledge accuracy (\%) on the available Multilingual MMLU languages for OLMo 3 7B. The Avg column includes English and the four evaluated target languages. \textbf{Bold} indicates the best score per column, and \underline{underline} indicates the second best.}
\label{tab:7b_mmlu}
\end{table*}
\begin{table*}[ht]
\centering
\begin{tabular}{lccccc}
\toprule
\textbf{Method} & \textbf{E$\to$X} & \textbf{X$\to$E} & \textbf{XQuAD} & \textbf{Belebele} & \textbf{M\_MMLU} \\
 & \textbf{chrF++} & \textbf{chrF++} & \textbf{EM (\%)} & \textbf{Acc. (\%)} & \textbf{Acc. (\%)} \\
\midrule
\multicolumn{6}{c}{\textbf{OLMo 2 1B}} \\
\midrule
Baseline & \underline{30.37} & 43.36 & 14.73 & 24.83 & 26.82 \\
VE + Mean Init & 28.06\textsuperscript{+} & \underline{47.58}\textsuperscript{*} & \underline{20.25}\textsuperscript{*} & 25.29 & 28.95\textsuperscript{*} \\
VE + FOCUS & 23.30\textsuperscript{+} & 45.43\textsuperscript{*} &  17.67\textsuperscript{*} & 25.21 & \textbf{30.08}\textsuperscript{*} \\
VE + ZeTT & 23.50\textsuperscript{+} & 39.99\textsuperscript{+} & 17.94\textsuperscript{*} & 24.42 & 28.23 \\
VE + Sakajo et al. & 28.39\textsuperscript{+} & 46.20\textsuperscript{*} & 15.17 & \underline{27.00}\textsuperscript{*} & 26.78 \\
VE + TokAlign & 25.23\textsuperscript{+} & 45.26\textsuperscript{*} & 16.85\textsuperscript{*} & 25.00 & 28.28\textsuperscript{*} \\
\textbf{MATE} & \textbf{33.65}\textsuperscript{*} & \textbf{47.88}\textsuperscript{*} & \textbf{29.89}\textsuperscript{*} & \textbf{27.51}\textsuperscript{*} & \underline{28.96}\textsuperscript{*} \\
\bottomrule
\end{tabular}
\caption{Downstream performance of OLMo 2 1B across translation and commonsense reasoning tasks. Bold and underline indicate the highest and second-highest values in each column. \textsuperscript{*} and \textsuperscript{+} indicate statistically significant improvement and degradation, respectively, relative to Baseline ($p<0.05$).}
\label{tab:olmo2_summary}
\end{table*}
\begin{table*}[ht]
  \centering
  \setlength{\tabcolsep}{3pt}
  \begin{tabular}{lccccccccccc}
  \toprule
   & \textbf{am} & \textbf{be} & \textbf{es} & \textbf{id} & \textbf{km} & \textbf{lo} & \textbf{my} & \textbf{ta} & \textbf{th} & \textbf{zh} & \textbf{Avg} \\
  \midrule
  \multicolumn{12}{c}{\textbf{English $\to$ Target}} \\
  \midrule
  Baseline & \underline{15.17} & 27.19 & \textbf{51.78} & 67.43 & \underline{17.43} & \underline{18.72} & \textbf{14.60} & 32.46 & 34.74 & 24.15 & \underline{30.37} \\
  VE + Mean Init & 12.56 & 23.29 & 50.82 & 67.24 & 12.15 & 13.01 & 9.74 & 33.52 & 35.29 & 23.02 & 28.06 \\
  VE + FOCUS & 5.81 & 17.24 & 49.70 & 66.62 & 10.76 & 9.02 & 8.95 & 18.84 & 27.05 & 19.01 & 23.30 \\
  VE + ZeTT & 7.12 & 12.67 & 49.78 & 67.15 & 9.13 & 11.11 & 10.14 & 19.45 & 24.80 & 23.69 & 23.50 \\
  VE + Sakajo et al. & 8.08 & 23.66 & \underline{51.43} & \underline{67.59} & 12.77 & 15.12 & \underline{11.61} & \underline{35.99} & \underline{35.31} & 22.33 & 28.39 \\
  VE + TokAlign & 9.18 & \underline{28.67} & 50.48 & 67.57 & 10.47 & 12.06 & 8.94 & 18.54 & 21.25 & \textbf{25.13} & 25.23 \\
  \textbf{MATE} & \textbf{16.77} & \textbf{30.86} & 49.08 & \textbf{68.28} & \textbf{20.47} & \textbf{30.19} & 11.54 & \textbf{42.42} & \textbf{41.87} & \underline{25.06} & \textbf{33.65} \\
  \midrule
  \multicolumn{12}{c}{\textbf{Target $\to$ English}} \\
  \midrule
  Baseline & 44.38 & 42.85 & 51.50 & 62.83 & 30.12 & 36.33 & 23.70 & 40.18 & 50.32 & 51.41 & 43.36 \\
  VE + Mean Init & 44.83 & 46.72 & \textbf{55.60} & \underline{64.55} & \underline{42.53} & \underline{41.58} & 28.05 & \textbf{48.14} & 52.16 & \underline{51.64} & \underline{47.58} \\
  VE + FOCUS & 43.28 & 44.53 & 53.00 & 61.63 & 37.32 & 40.11 & \underline{29.37} & 45.01 & 51.25 & 48.78 & 45.43 \\
  VE + ZeTT & 33.85 & 41.50 & 51.33 & 61.89 & 27.11 & 28.77 & 21.83 & 34.65 & 49.22 & 49.73 & 39.99 \\
  VE + Sakajo et al. & \underline{45.33} & \textbf{47.32} & 54.53 & 64.07 & 34.24 & 36.83 & \textbf{30.05} & 45.40 & \underline{52.35} & \textbf{51.84} & 46.20 \\
  VE + TokAlign & 43.49 & \underline{46.93} & \underline{54.75} & 63.96 & 33.43 & 37.55 & 26.75 & 42.53 & 52.23 & 50.97 & 45.26 \\
  \textbf{MATE} & \textbf{45.66} & 46.02 & 51.36 & \textbf{64.58} & \textbf{43.45} & \textbf{48.58} & 26.53 & \underline{47.87} & \textbf{53.46} & 51.33 & \textbf{47.88} \\
  \bottomrule
  \end{tabular}
  \caption{Language-specific OLMo 2 1B translation chrF++ scores on FLORES-200. \textbf{Bold} indicates the best score per column, and \underline{underline} indicates the second best.}
  \label{tab:main_translation_chrf}
\end{table*}
\begin{table*}[ht]
  \centering
  \begin{tabular*}{\textwidth}{@{\extracolsep{\fill}}lccccc@{}}
  \toprule
  \textbf{Method} & \textbf{en} & \textbf{es} & \textbf{th} & \textbf{zh} & \textbf{Avg} \\
  \midrule
  Baseline & 45.80 & 2.52 & 1.01 & 9.58 & 14.73 \\
  VE + Mean Init & \underline{51.51} & \underline{14.79} & 1.76 & \underline{12.94} & \underline{20.25} \\
  VE + FOCUS & 49.33 & 12.77 & 1.60 & 6.97 & 17.67 \\
  VE + ZeTT & \textbf{52.44} & 6.89 & 1.60 & 10.84 & 17.94 \\
  VE + Sakajo et al. & 49.24 & 4.03 & 1.51 & 5.88 & 15.17 \\
  VE + TokAlign & 49.24 & 5.63 & \underline{2.18} & 10.34 & 16.85 \\
  \textbf{MATE} & 51.43 & \textbf{22.44} & \textbf{20.00} & \textbf{25.71} & \textbf{29.89} \\
  \bottomrule
  \end{tabular*}
  \caption{Language-specific OLMo 2 1B XQuAD exact match scores (\%). \textbf{Bold} indicates the best score per column, and \underline{underline} indicates the second best.}
  \label{tab:main_xquad}
\end{table*}
\begin{table*}[ht]
  \centering
  \setlength{\tabcolsep}{2.5pt}
  \begin{tabular}{lcccccccccccc}
  \toprule
  \textbf{Method} & \textbf{en} & \textbf{am} & \textbf{es} & \textbf{id} & \textbf{km} & \textbf{lo} & \textbf{my} & \textbf{ta} & \textbf{th} & \textbf{zh} & \textbf{Avg} \\
  \midrule
  Baseline & 30.22 & 23.11 & \underline{26.44} & 27.00 & 21.67 & 20.89 & 24.22 & 23.33 & 26.11 & 25.33 & 24.83 \\
  VE + Mean Init & 29.22 & \underline{25.11} & 23.56 & 23.78 & 25.22 & 20.78 & \textbf{27.22} & 25.89 & 24.00 & 28.11 & 25.29 \\
  VE + FOCUS & \textbf{30.89} & 24.11 & 25.89 & 26.78 & 21.78 & 22.44 & 26.22 & 22.44 & 22.89 & \underline{28.67} & 25.21 \\
  VE + ZeTT & 28.22 & 23.33 & 25.89 & 25.78 & 20.78 & 21.33 & 24.56 & 25.22 & 22.56 & 26.56 & 24.42 \\
  VE + Sakajo et al. & 28.56 & 23.22 & 26.00 & \underline{27.67} & \underline{26.56} & \textbf{26.78} & \underline{26.33} & \textbf{28.78} & \underline{27.33} & \textbf{28.78} & \underline{27.00} \\
  VE + TokAlign & 26.56 & \textbf{25.89} & 23.44 & 27.11 & 24.56 & 25.00 & 25.11 & 21.67 & 23.78 & 26.89 & 25.00 \\
  \textbf{MATE} & \underline{30.44} & 24.33 & \textbf{28.44} & \textbf{29.22} & \textbf{27.00} & \underline{26.44} & 24.11 & \underline{28.44} & \textbf{28.33} & 28.33 & \textbf{27.51} \\
  \bottomrule
  \end{tabular}
  \caption{Language-specific OLMo 2 1B Belebele reading comprehension accuracy (\%). \textbf{Bold} indicates the best score per column, and \underline{underline} indicates the second best.}
  \label{tab:main_belebele}
\end{table*}
\begin{table*}[ht]
  \centering
  \begin{tabular*}{\textwidth}{@{\extracolsep{\fill}}lcccccc@{}}
  \toprule
  \textbf{Method} & \textbf{en} & \textbf{es} & \textbf{id} & \textbf{ta} & \textbf{zh} & \textbf{Avg} \\
  \midrule
  Baseline & 30.06 & 26.07 & 26.86 & 25.11 & 25.99 & 26.82 \\
  VE + Mean Init & 33.93 & \underline{29.79} & 28.12 & \underline{25.48} & \underline{27.45} & 28.95 \\
  VE + FOCUS & \textbf{34.99} & \textbf{30.25} & \underline{29.88} & \textbf{26.41} & \textbf{28.87} & \textbf{30.08} \\
  VE + ZeTT & 33.50 & 28.15 & 28.89 & 24.49 & 26.12 & 28.23 \\
  VE + Sakajo et al. & 30.03 & 26.37 & 26.46 & 24.94 & 26.09 & 26.78 \\
  VE + TokAlign & \underline{34.33} & 28.98 & 28.05 & 23.31 & 26.71 & 28.28 \\
  \textbf{MATE} & 32.98 & 29.26 & \textbf{30.28} & 25.10 & 27.18 & \underline{28.96} \\
  \bottomrule
  \end{tabular*}
  \caption{Language-specific OLMo 2 1B Multilingual MMLU accuracy (\%). \textbf{Bold} indicates the best score per column, and \underline{underline} indicates the second best.}
  \label{tab:main_mmlu}
\end{table*}
\begin{table*}[ht]
\centering
\begin{tabular}{lccccc}
\toprule
\textbf{Method} & \textbf{E$\to$X} & \textbf{X$\to$E} & \textbf{XQuAD} & \textbf{Belebele} & \textbf{M\_MMLU} \\
 & \textbf{chrF++} & \textbf{chrF++} & \textbf{EM (\%)} & \textbf{Acc. (\%)} & \textbf{Acc. (\%)} \\
\midrule
\multicolumn{6}{c}{\textbf{Pythia 1B}} \\
\midrule
Baseline & \underline{36.60} & 44.27 & \underline{7.29} & \underline{25.64} & \textbf{24.95} \\
VE + Mean Init & 35.57\textsuperscript{+} & 44.49 & 4.71\textsuperscript{+} & 24.61 & 24.58 \\
VE + FOCUS & 35.46\textsuperscript{+} & 44.46 & 3.91\textsuperscript{+} & 25.09 & 24.74 \\
VE + ZeTT & 35.65\textsuperscript{+} & 44.40 & 4.79\textsuperscript{+} & 25.24 & \underline{24.82} \\
VE + Sakajo et al. & 35.80\textsuperscript{+} & \underline{45.09}\textsuperscript{*} & 3.68\textsuperscript{+} & 25.57 & 23.97\textsuperscript{+} \\
VE + TokAlign & 35.16\textsuperscript{+} & 44.20 & 3.05\textsuperscript{+} & \textbf{26.10} & 24.43\textsuperscript{+} \\
\textbf{MATE} & \textbf{37.55}\textsuperscript{*} & \textbf{47.03}\textsuperscript{*} & \textbf{8.26}\textsuperscript{*} & 25.07 & 24.58 \\
\bottomrule
\end{tabular}
\caption{Downstream performance of Pythia 1B across translation and commonsense reasoning tasks. Bold and underline indicate the highest and second-highest values in each column. \textsuperscript{*} and \textsuperscript{+} indicate statistically significant improvement and degradation, respectively, relative to Baseline ($p<0.05$).}
\label{tab:pythia_summary}
\end{table*}
\begin{table*}[ht]
  \centering
  \setlength{\tabcolsep}{3pt}
  \begin{tabular}{lccccccccccc}
  \toprule
   & \textbf{am} & \textbf{be} & \textbf{es} & \textbf{id} & \textbf{km} & \textbf{lo} & \textbf{my} & \textbf{ta} & \textbf{th} & \textbf{zh} & \textbf{Avg} \\
  \midrule
  \multicolumn{12}{c}{\textbf{English $\to$ Target}} \\
  \midrule
  Baseline & \textbf{28.84} & \underline{37.07} & 51.21 & 67.24 & \underline{21.23} & 28.69 & \textbf{19.32} & \underline{46.28} & \underline{41.17} & 24.98 & \underline{36.60} \\
  VE + Mean Init & 24.96 & 36.97 & 51.24 & 67.43 & 19.68 & 27.70 & 16.78 & 45.98 & 39.94 & 24.99 & 35.57 \\
  VE + FOCUS & 25.30 & \textbf{37.11} & \underline{51.45} & \textbf{67.64} & 18.77 & 26.72 & 16.55 & 45.78 & 40.19 & 25.13 & 35.46 \\
  VE + ZeTT & 25.30 & 36.77 & \textbf{51.51} & 67.28 & 18.96 & \underline{29.32} & 16.57 & 45.43 & 39.97 & \textbf{25.41} & 35.65 \\
  VE + Sakajo et al. & 25.58 & 36.83 & 51.18 & \underline{67.50} & 18.06 & 28.68 & \underline{18.31} & \textbf{46.78} & 40.55 & 24.57 & 35.80 \\
  VE + TokAlign & 23.54 & 36.93 & 51.37 & 67.33 & 18.94 & 26.86 & 16.86 & 44.89 & 39.94 & 24.96 & 35.16 \\
  \textbf{MATE} & \underline{28.75} & 36.35 & 48.48 & \textbf{67.64} & \textbf{26.42} & \textbf{36.32} & 18.10 & 45.52 & \textbf{42.78} & \underline{25.14} & \textbf{37.55} \\
  \midrule
  \multicolumn{12}{c}{\textbf{Target $\to$ English}} \\
  \midrule
  Baseline & 41.08 & 46.03 & 54.43 & 63.59 & 28.55 & 34.64 & 28.81 & 45.04 & \underline{49.97} & 50.52 & 44.27 \\
  VE + Mean Init & 43.41 & 45.73 & \underline{54.50} & 63.78 & 30.24 & 33.42 & 28.89 & 44.26 & 49.36 & \underline{51.30} & 44.49 \\
  VE + FOCUS & 42.49 & 45.37 & \textbf{54.62} & 63.60 & \underline{30.73} & 32.54 & \underline{29.80} & \underline{45.57} & 49.54 & 50.37 & 44.46 \\
  VE + ZeTT & \underline{43.75} & 45.68 & 54.36 & 63.50 & 30.24 & 34.65 & 28.33 & 43.75 & 49.41 & 50.37 & 44.40 \\
  VE + Sakajo et al. & \textbf{44.32} & \textbf{46.13} & 54.45 & 63.82 & 29.95 & \underline{35.31} & \textbf{30.98} & 44.78 & 49.70 & \textbf{51.44} & \underline{45.09} \\
  VE + TokAlign & 43.30 & \underline{46.07} & 54.45 & \underline{63.84} & 28.82 & 31.93 & 29.49 & 43.88 & 49.64 & 50.62 & 44.20 \\
  \textbf{MATE} & 43.66 & 45.86 & 54.29 & \textbf{63.91} & \textbf{39.13} & \textbf{47.09} & 26.90 & \textbf{45.94} & \textbf{52.47} & 51.06 & \textbf{47.03} \\
  \bottomrule
  \end{tabular}
  \caption{Language-specific Pythia 1B translation chrF++ scores on FLORES-200. \textbf{Bold} indicates the best score per column, and \underline{underline} indicates the second best.}
  \label{tab:main_translation_chrf_pythia}
\end{table*}
\begin{table*}[ht]
  \centering
  \begin{tabular*}{\textwidth}{@{\extracolsep{\fill}}lccccc@{}}
  \toprule
  \textbf{Method} & \textbf{en} & \textbf{es} & \textbf{th} & \textbf{zh} & \textbf{Avg} \\
  \midrule
  Baseline & \textbf{27.48} & 0.50 & 0.00 & 1.18 & \underline{7.29} \\
  VE + Mean Init & 15.63 & 0.50 & \underline{1.01} & \underline{1.68} & 4.71 \\
  VE + FOCUS & 13.53 & 0.84 & 0.76 & 0.50 & 3.91 \\
  VE + ZeTT & \underline{18.07} & 0.42 & 0.59 & 0.08 & 4.79 \\
  VE + Sakajo et al. & 13.45 & \underline{0.92} & 0.00 & 0.34 & 3.68 \\
  VE + TokAlign & 11.43 & 0.17 & 0.25 & 0.34 & 3.05 \\
  \textbf{MATE} & 17.31 & \textbf{10.17} & \textbf{2.10} & \textbf{3.45} & \textbf{8.26} \\
  \bottomrule
  \end{tabular*}
  \caption{Language-specific Pythia 1B XQuAD exact match scores (\%). \textbf{Bold} indicates the best score per column, and \underline{underline} indicates the second best.}
  \label{tab:main_xquad_pythia}
\end{table*}
\begin{table*}[ht]
  \centering
  \setlength{\tabcolsep}{2.5pt}
  \begin{tabular}{lccccccccccc}
  \toprule
  \textbf{Method} & \textbf{en} & \textbf{am} & \textbf{es} & \textbf{id} & \textbf{km} & \textbf{lo} & \textbf{my} & \textbf{ta} & \textbf{th} & \textbf{zh} & \textbf{Avg} \\
  \midrule
  Baseline & 24.89 & 24.33 & \textbf{27.00} & \textbf{28.78} & 24.33 & 23.22 & 24.56 & 26.00 & \underline{26.44} & \underline{26.89} & \underline{25.64} \\
  VE + Mean Init & 22.56 & 24.00 & 25.56 & 25.00 & \underline{27.22} & 24.67 & 22.89 & 23.78 & 23.78 & 26.67 & 24.61 \\
  VE + FOCUS & 23.78 & 24.00 & \textbf{27.00} & 24.67 & 26.44 & 24.11 & 21.78 & 25.67 & 25.67 & \textbf{27.78} & 25.09 \\
  VE + ZeTT & \textbf{26.33} & 23.22 & 25.56 & 24.22 & 25.00 & 23.22 & 24.56 & 24.22 & \textbf{29.33} & 26.78 & 25.24 \\
  VE + Sakajo et al. & 24.22 & \underline{25.78} & 23.33 & 25.44 & \textbf{27.67} & \textbf{27.22} & 23.33 & \textbf{26.89} & 25.11 & 26.67 & 25.57 \\
  VE + TokAlign & 23.56 & \textbf{28.00} & \underline{26.22} & \underline{26.78} & 26.00 & 24.67 & \underline{26.67} & 25.89 & 26.33 & \underline{26.89} & \textbf{26.10} \\
  \textbf{MATE} & \underline{25.11} & 23.67 & 24.33 & 23.89 & 26.44 & \underline{24.78} & \textbf{27.11} & \underline{26.11} & 24.44 & 24.78 & 25.07 \\
  \bottomrule
  \end{tabular}
  \caption{Language-specific Pythia 1B Belebele reading comprehension accuracy (\%). \textbf{Bold} indicates the best score per column, and \underline{underline} indicates the second best.}
  \label{tab:main_belebele_pythia}
\end{table*}
\begin{table*}[ht]
  \centering
  \begin{tabular*}{\textwidth}{@{\extracolsep{\fill}}lcccccc@{}}
  \toprule
  \textbf{Method} & \textbf{en} & \textbf{es} & \textbf{id} & \textbf{ta} & \textbf{zh} & \textbf{Avg} \\
  \midrule
  Baseline & \textbf{24.68} & 25.23 & 25.19 & \textbf{24.89} & 24.77 & \textbf{24.95} \\
  VE + Mean Init & 24.49 & 25.36 & 25.49 & 23.40 & 24.15 & 24.58 \\
  VE + FOCUS & 23.97 & \underline{25.39} & 25.39 & \underline{24.43} & 24.51 & 24.74 \\
  VE + ZeTT & \underline{24.53} & 25.12 & \underline{25.50} & 23.98 & \underline{24.94} & \underline{24.82} \\
  VE + Sakajo et al. & 23.40 & 24.75 & \textbf{25.75} & 22.71 & 23.21 & 23.97 \\
  VE + TokAlign & 23.55 & \textbf{25.42} & 24.71 & 23.52 & \textbf{24.98} & 24.43 \\
  \textbf{MATE} & 24.34 & 25.33 & 25.28 & 24.12 & 23.81 & 24.58 \\
  \bottomrule
  \end{tabular*}
  \caption{Language-specific Pythia 1B Multilingual MMLU accuracy (\%). \textbf{Bold} indicates the best score per column, and \underline{underline} indicates the second best.}
\label{tab:main_mmlu_pythia}
\end{table*}
\begin{table}[ht]
\centering
\begin{tabular}{lccc}
\toprule
\textbf{Hyperparameter} & \textbf{OLMo 2 1B} & \textbf{Pythia 1B} & \textbf{OLMo 3 7B} \\
\midrule
Global batch size & 512 & 512 & 256 \\
Learning rate & $2.0 \times 10^{-4}$ & $2.0 \times 10^{-4}$ & $2.1 \times 10^{-4}$ \\
Weight decay & 0.1 & 0.1 & 0.1 \\
Optimizer & AdamW & AdamW & AdamW \\
($\beta_1$, $\beta_2$) & (0.9, 0.95) & (0.9, 0.95) & (0.9, 0.95) \\
Gradient clip & 1.0 & 1.0 & 1.0 \\
Max sequence length & 4096 & 2048 & 8192 \\
\bottomrule
\end{tabular}
\caption{Training hyperparameters for OLMo 2 1B, Pythia 1B, and OLMo 3 7B.}
\label{tab:training_hyperparameters}
\end{table}
\begin{table*}[ht]
\centering
\setlength{\tabcolsep}{2.5pt}
\begin{tabular}{lccccccccccc}
\toprule
\textbf{Method} & \textbf{am} & \textbf{be} & \textbf{es} & \textbf{id} & \textbf{km} & \textbf{lo} & \textbf{my} & \textbf{ta} & \textbf{th} & \textbf{zh} & \textbf{Avg} \\
\midrule
\multicolumn{12}{c}{\textbf{English $\to$ Target}} \\
\midrule
0\% & 19.20 & 30.08 & 50.05 & 68.82 & \textbf{25.31} & 31.27 & \textbf{18.96} & 48.78 & 46.17 & \underline{27.85} & 36.65 \\
12.5\% & 16.92 & 30.69 & \underline{50.34} & \textbf{68.98} & 23.53 & 31.10 & 17.19 & 49.61 & \underline{47.12} & 27.71 & 36.32 \\
43.75\% & \underline{21.82} & \underline{33.28} & \textbf{50.41} & \underline{68.83} & 24.81 & \textbf{34.19} & 17.45 & \textbf{50.05} & 46.35 & \textbf{28.40} & \underline{37.56} \\
87.5\% & \textbf{22.79} & \textbf{33.77} & 50.10 & 68.04 & \underline{24.92} & \underline{33.65} & \underline{18.46} & \underline{49.68} & \textbf{47.39} & 27.44 & \textbf{37.62} \\
93.75\% & 12.79 & 24.25 & 50.12 & 67.74 & 21.83 & 31.19 & 17.28 & 48.41 & 46.18 & 26.60 & 34.64 \\
\midrule
\multicolumn{12}{c}{\textbf{Target $\to$ English}} \\
\midrule
0\% & \underline{53.36} & \underline{48.87} & \textbf{57.14} & \textbf{67.02} & \textbf{51.67} & \underline{53.93} & \textbf{37.46} & \underline{53.98} & \textbf{58.04} & \textbf{55.71} & \textbf{53.72} \\
12.5\% & \textbf{54.27} & 48.69 & \underline{57.01} & 66.54 & 49.59 & \textbf{53.95} & \underline{37.25} & \textbf{54.41} & 57.54 & 55.13 & \underline{53.44} \\
43.75\% & 53.33 & \textbf{49.17} & 55.65 & \underline{66.87} & 44.38 & 53.65 & 36.56 & 53.28 & 57.73 & \underline{55.27} & 52.59 \\
87.5\% & 52.51 & 48.33 & 57.00 & 66.37 & \underline{49.86} & 52.79 & 36.59 & 52.74 & \underline{57.85} & 54.98 & 52.90 \\
93.75\% & 52.43 & 48.38 & 55.71 & 65.89 & 46.46 & 50.81 & 36.48 & 51.92 & 56.92 & 54.70 & 51.97 \\
\bottomrule
\end{tabular}
\caption{Language-specific OLMo 3 7B translation chrF++ scores for embedding alignment ratio sensitivity. Averages cover the ten target languages in each direction. Bold marks the best score and underline the second best in each column.}
\label{tab:ratio_translation_chrf}
\end{table*}
\begin{table*}[ht]
\centering
\setlength{\tabcolsep}{4pt}
\begin{tabular}{lccccc}
\toprule
\textbf{Method} & \textbf{en} & \textbf{es} & \textbf{th} & \textbf{zh} & \textbf{Avg} \\
\midrule
0\% & \underline{67.65} & \underline{40.08} & 38.40 & 31.43 & 44.39 \\
12.5\% & \textbf{67.90} & \textbf{42.18} & \textbf{39.58} & 32.86 & \underline{45.63} \\
43.75\% & 63.03 & 36.72 & 29.75 & 36.13 & 41.41 \\
87.5\% & 65.88 & 38.57 & 37.06 & \textbf{38.32} & 44.96 \\
93.75\% & 66.13 & 39.58 & \underline{39.41} & \underline{37.73} & \textbf{45.71} \\
\bottomrule
\end{tabular}
\caption{Language-specific OLMo 3 7B XQuAD exact match scores for embedding alignment ratio sensitivity. Bold marks the best score and underline the second best in each column.}
\label{tab:ratio_xquad}
\end{table*}
\begin{table*}[ht]
\centering
\setlength{\tabcolsep}{2.5pt}
\begin{tabular}{lccccccccccc}
\toprule
\textbf{Method} & \textbf{en} & \textbf{am} & \textbf{es} & \textbf{id} & \textbf{km} & \textbf{lo} & \textbf{my} & \textbf{ta} & \textbf{th} & \textbf{zh} & \textbf{Avg} \\
\midrule
0\% & \underline{69.22} & 29.22 & \textbf{55.11} & \textbf{54.11} & \underline{30.44} & 33.22 & \textbf{30.11} & \underline{37.78} & \textbf{43.56} & \underline{55.22} & \textbf{43.80} \\
12.5\% & 62.33 & 29.11 & 43.78 & 46.78 & 26.67 & 29.56 & 26.00 & 35.00 & 34.78 & 53.78 & 38.78 \\
43.75\% & \textbf{69.89} & \textbf{31.78} & \underline{52.56} & \underline{53.44} & 28.78 & \underline{34.44} & \underline{26.67} & \textbf{39.56} & 39.67 & \textbf{55.67} & \underline{43.24} \\
87.5\% & 63.44 & 30.22 & 51.22 & 50.22 & 30.11 & \textbf{34.67} & 26.56 & 34.78 & \underline{40.56} & 48.56 & 41.03 \\
93.75\% & 61.67 & \underline{30.44} & 49.56 & 50.44 & \textbf{31.00} & 32.78 & 25.00 & 34.67 & 36.67 & 52.44 & 40.47 \\
\bottomrule
\end{tabular}
\caption{Language-specific OLMo 3 7B Belebele accuracy scores for embedding alignment ratio sensitivity. Bold marks the best score and underline the second best in each column.}
\label{tab:ratio_belebele}
\end{table*}
\begin{table*}[ht]
\centering
\setlength{\tabcolsep}{4pt}
\begin{tabular}{lcccccc}
\toprule
\textbf{Method} & \textbf{en} & \textbf{es} & \textbf{id} & \textbf{ta} & \textbf{zh} & \textbf{Avg} \\
\midrule
0\% & 53.35 & \textbf{44.98} & 43.72 & \textbf{30.51} & \underline{39.54} & \underline{42.42} \\
12.5\% & 52.90 & \underline{44.47} & \underline{43.85} & \underline{30.09} & 39.09 & 42.08 \\
43.75\% & \underline{53.64} & 44.30 & \textbf{43.98} & 30.00 & \textbf{40.39} & \textbf{42.46} \\
87.5\% & 52.32 & 44.09 & 42.96 & 29.04 & 36.26 & 40.93 \\
93.75\% & \textbf{53.67} & 44.24 & 43.17 & 29.76 & 38.66 & 41.90 \\
\bottomrule
\end{tabular}
\caption{Language-specific OLMo 3 7B Multilingual MMLU accuracy scores for embedding alignment ratio sensitivity. Bold marks the best score and underline the second best in each column.}
\label{tab:ratio_mmlu}
\end{table*}
\begin{table*}[ht]
\centering
\setlength{\tabcolsep}{2.5pt}
\begin{tabular}{lccccccccccc}
\toprule
\textbf{Method} & \textbf{am} & \textbf{be} & \textbf{es} & \textbf{id} & \textbf{km} & \textbf{lo} & \textbf{my} & \textbf{ta} & \textbf{th} & \textbf{zh} & \textbf{Avg} \\
\midrule
\multicolumn{12}{c}{\textbf{English $\to$ Target}} \\
\midrule
Baseline & \textbf{30.05} & 33.15 & \textbf{52.75} & 68.12 & 23.28 & 26.62 & \underline{21.04} & \textbf{50.41} & 46.64 & 27.25 & \underline{37.93} \\
+PVE & \underline{26.09} & \textbf{34.46} & 49.80 & \underline{68.44} & 23.43 & 31.04 & \textbf{22.12} & 49.62 & \underline{47.03} & \underline{27.62} & \textbf{37.97} \\
+PVE+SEI & 19.20 & 30.08 & 50.05 & \textbf{68.82} & \textbf{25.31} & \underline{31.27} & 18.96 & 48.78 & 46.17 & \textbf{27.85} & 36.65 \\
\textbf{MATE} & 22.79 & \underline{33.77} & \underline{50.10} & 68.04 & \underline{24.92} & \textbf{33.65} & 18.46 & \underline{49.68} & \textbf{47.39} & 27.44 & 37.62 \\
\midrule
\multicolumn{12}{c}{\textbf{Target $\to$ English}} \\
\midrule
Baseline & 48.64 & 47.84 & 56.43 & 65.79 & 43.57 & 47.46 & \underline{36.69} & 43.62 & 57.75 & 54.62 & 50.24 \\
+PVE & 52.07 & 47.92 & 53.07 & 65.98 & 46.54 & 49.83 & 36.57 & \underline{53.02} & 57.22 & 52.94 & 51.52 \\
+PVE+SEI & \textbf{53.36} & \textbf{48.87} & \textbf{57.14} & \textbf{67.02} & \textbf{51.67} & \textbf{53.93} & \textbf{37.46} & \textbf{53.98} & \textbf{58.04} & \textbf{55.71} & \textbf{53.72} \\
\textbf{MATE} & \underline{52.51} & \underline{48.33} & \underline{57.00} & \underline{66.37} & \underline{49.86} & \underline{52.79} & 36.59 & 52.74 & \underline{57.85} & \underline{54.98} & \underline{52.90} \\
\bottomrule
\end{tabular}
\caption{Language-specific OLMo 3 7B translation chrF++ scores for the component ablation. Averages cover the ten target languages in each direction. Bold marks the best score and underline the second best in each column.}
\label{tab:translation_chrf}
\end{table*}
\begin{table*}[ht]
\centering
\setlength{\tabcolsep}{4pt}
\begin{tabular}{lccccc}
\toprule
\textbf{Method} & \textbf{en} & \textbf{es} & \textbf{th} & \textbf{zh} & \textbf{Avg} \\
\midrule
Baseline & \underline{65.97} & 33.36 & 1.68 & 21.60 & 30.65 \\
+PVE & 62.86 & 20.59 & 17.73 & 17.56 & 29.68 \\
+PVE+SEI & \textbf{67.65} & \textbf{40.08} & \textbf{38.40} & \underline{31.43} & \underline{44.39} \\
\textbf{MATE} & 65.88 & \underline{38.57} & \underline{37.06} & \textbf{38.32} & \textbf{44.96} \\
\bottomrule
\end{tabular}
\caption{Language-specific OLMo 3 7B XQuAD exact match scores for the component ablation. Bold marks the best score and underline the second best in each column.}
\label{tab:ablation_xquad}
\end{table*}
\begin{table*}[ht]
\centering
\setlength{\tabcolsep}{2.5pt}
\begin{tabular}{lccccccccccc}
\toprule
\textbf{Method} & \textbf{en} & \textbf{am} & \textbf{es} & \textbf{id} & \textbf{km} & \textbf{lo} & \textbf{my} & \textbf{ta} & \textbf{th} & \textbf{zh} & \textbf{Avg} \\
\midrule
Baseline & \underline{66.56} & 27.89 & 50.22 & 44.78 & 28.56 & 28.44 & 27.33 & 32.56 & 34.33 & \underline{54.78} & 39.54 \\
+PVE & 64.33 & 28.11 & \underline{52.44} & \underline{50.89} & \textbf{31.89} & 29.33 & \underline{27.89} & 31.67 & \underline{40.67} & 53.78 & \underline{41.10} \\
+PVE+SEI & \textbf{69.22} & \underline{29.22} & \textbf{55.11} & \textbf{54.11} & \underline{30.44} & \underline{33.22} & \textbf{30.11} & \textbf{37.78} & \textbf{43.56} & \textbf{55.22} & \textbf{43.80} \\
\textbf{MATE} & 63.44 & \textbf{30.22} & 51.22 & 50.22 & 30.11 & \textbf{34.67} & 26.56 & \underline{34.78} & 40.56 & 48.56 & 41.03 \\
\bottomrule
\end{tabular}
\caption{Language-specific OLMo 3 7B Belebele accuracy scores for the component ablation. Bold marks the best score and underline the second best in each column.}
\label{tab:belebele}
\end{table*}
\begin{table*}[ht]
\centering
\setlength{\tabcolsep}{4pt}
\begin{tabular}{lcccccc}
\toprule
\textbf{Method} & \textbf{en} & \textbf{es} & \textbf{id} & \textbf{ta} & \textbf{zh} & \textbf{Avg} \\
\midrule
Baseline & 50.66 & 40.13 & 39.26 & 26.75 & 36.80 & 38.72 \\
+PVE & \textbf{53.97} & 42.40 & 42.00 & \underline{29.71} & \underline{38.21} & \underline{41.26} \\
+PVE+SEI & \underline{53.35} & \textbf{44.98} & \textbf{43.72} & \textbf{30.51} & \textbf{39.54} & \textbf{42.42} \\
\textbf{MATE} & 52.32 & \underline{44.09} & \underline{42.96} & 29.04 & 36.26 & 40.93 \\
\bottomrule
\end{tabular}
\caption{Language-specific OLMo 3 7B Multilingual MMLU accuracy scores for the component ablation. Bold marks the best score and underline the second best in each column.}
\label{tab:m_mmlu}
\end{table*}

\end{document}